\documentclass{article}

\usepackage{iclr2027_conference,times}

\usepackage{amsmath,amsfonts,bm}

\def\eqref#1{equation~\ref{#1}}

\def\1{\bm{1}}

\DeclareMathAlphabet{\mathsfit}{\encodingdefault}{\sfdefault}{m}{sl}
\SetMathAlphabet{\mathsfit}{bold}{\encodingdefault}{\sfdefault}{bx}{n}

\definecolor{rglink}{RGB}{40,74,120}
\definecolor{rgpink}{RGB}{214,51,132}   
\definecolor{rgpinkbg}{RGB}{253,231,243} 
\usepackage{hyperref}
\hypersetup{
  colorlinks=true,
  linkcolor=rglink,
  citecolor=rglink,
  urlcolor=rglink
}
\usepackage{url}
\usepackage{xurl}

\usepackage{amsmath}
\usepackage{amssymb}
\usepackage{graphicx}
\usepackage{booktabs}
\usepackage{multirow}
\usepackage{array}
\usepackage{colortbl}
\usepackage{placeins}
\usepackage{needspace}
\usepackage{microtype}
\usepackage{subcaption}  
\usepackage{tcolorbox}
\usepackage{xcolor}
\usepackage{soul}
\usepackage{wrapfig}
\usepackage{algorithm}
\usepackage{algpseudocode}

\algrenewcommand{\algorithmicrequire}{\textbf{Input:}}
\algrenewcommand{\algorithmicensure}{\textbf{Output:}}
\newif\ifshowchanges \showchangestrue
\definecolor{prevedit}{RGB}{0,84,180}   
\definecolor{newedit}{RGB}{190,0,110}   
\DeclareRobustCommand{\prevedit}[1]{\ifshowchanges\ifvmode\leavevmode\fi{\color{prevedit}#1}\else#1\fi}
\DeclareRobustCommand{\newedit}[1]{\ifshowchanges\ifvmode\leavevmode\fi{\color{newedit}#1}\else#1\fi}
\providecommand{\acknowledgments}[1]{}
\newenvironment{contribbox}
  {\begin{tcolorbox}[colback=blue!5!white, colframe=blue!60!black,
    before skip=0.25cm, after skip=0.25cm, boxsep=0.05cm,
    left=0.2cm, right=0.2cm, top=0.1cm, bottom=0.1cm]}
  {\end{tcolorbox}}

\definecolor{rgrow}{gray}{0.92}
\definecolor{rgaccent}{RGB}{232,240,255}
\definecolor{rgsummary}{RGB}{238,242,246}
\definecolor{rgdrop}{RGB}{255,247,226}
\definecolor{rgbest}{RGB}{190,214,236}   
\definecolor{rgopen}{RGB}{247,230,208}   
\definecolor{mgold}{RGB}{201,162,39}
\definecolor{msilver}{RGB}{140,148,158}
\definecolor{mbronze}{RGB}{176,122,74}
\newcommand{\gplus}{\mbox{GLiDE\textsuperscript{+}}}
\newcommand{\mgd}{\textcolor{mgold}{$\blacklozenge$}}
\newcommand{\msv}{\textcolor{msilver}{$\blacklozenge$}}
\newcommand{\mbz}{\textcolor{mbronze}{$\blacklozenge$}}
\usepackage{pifont}
\usepackage{tikz}
\definecolor{rgyes}{RGB}{0,128,64}     
\definecolor{rgpart}{RGB}{215,130,0}   
\definecolor{rgno}{RGB}{175,175,175}   
\newcommand{\cmark}{\textcolor{rgyes}{\ding{51}}}
\newcommand{\xmark}{\textcolor{rgno}{\ding{55}}}
\newcommand{\pmark}{\tikz[baseline=-0.55ex]{\draw[rgpart,line width=0.5pt] (0,0) circle (0.75ex); \fill[rgpart] (0,-0.75ex) arc (-90:90:0.75ex) -- cycle;}}
\newsavebox{\keystatsbox}
\newsavebox{\compcapbox}
\newcommand{\rggobble}[1]{}
\makeatletter
\newcommand{\rgtablefont}{\fontsize{7}{8.5}\selectfont}
\newcommand{\rgmatchstretch}[4]{\begingroup\rgtablefont
  \dimen@=\dimexpr\ht#1+\dp#1-\ht#2-\dp#2\relax
  \ifdim\dimen@<\z@ \dimen@=\z@ \fi
  \count@=\numexpr#3*\baselineskip\relax
  \dimen@=\dimexpr#4pt+\dimen@*65536/\count@\relax
  \xdef\rgstretch{\strip@pt\dimen@}\endgroup}
\makeatother

\title{RoboGaze: Evaluating Robot World Models via Structured Vision-Language Analysis
}

\author{%
\begin{minipage}{\dimexpr\textwidth-2\tabcolsep\relax}
\centering
\fontsize{10}{14}\selectfont\bfseries
Minh-Loi Nguyen\textsuperscript{1}\quad
Nghiem T. Diep\textsuperscript{2}\quad
Hung K. Nguyen\textsuperscript{2}\quad
Minh Le\textsuperscript{2}\\
Doanh L. Thien\textsuperscript{1,2}\quad
Tuan A. Tran\textsuperscript{2}\quad
Hoang H. Tran\textsuperscript{1}\quad
Dung D. Le\textsuperscript{1}\\
Vu N. Duong\textsuperscript{1}\quad
Khoa D. Doan\textsuperscript{2}\quad
Ngan Le\textsuperscript{7}\quad
Daniel Sonntag\textsuperscript{3}\\
An T. Le\textsuperscript{1,2,6}\quad
Vien Anh Ngo\textsuperscript{1,2}\quad
Duy M. H. Nguyen\textsuperscript{\ensuremath{\dagger},3,4,5}\quad
Tran V. Nhiem\textsuperscript{\ensuremath{\dagger},2}\\[10pt]
\normalfont\footnotesize
\textsuperscript{1}Center for AI Research, VinUniversity\\
\textsuperscript{2}VinRobotics\quad
\textsuperscript{3}DFKI\quad
\textsuperscript{4}University of Stuttgart\\
\textsuperscript{5}Max Planck Research School for Intelligent Systems\\
\textsuperscript{6}Technische Universit\"at Darmstadt\\
\textsuperscript{7}University of Arkansas, USA\\
\textsuperscript{\ensuremath{\dagger}}Project Leads.
\end{minipage}%
}

\iclrfinalcopy
\showchangesfalse
\hypersetup{pdfauthor={Minh-Loi Nguyen, Nghiem T. Diep, Hung K. Nguyen, Minh Le, Doanh L. Thien, Tuan A. Tran, Hoang H. Tran, Dung D. Le, Vu N. Duong, Khoa D. Doan, Ngan Le, Daniel Sonntag, An T. Le, Vien Anh Ngo, Duy M. H. Nguyen, Tran V. Nhiem}}

\begin{document}
\raggedbottom
\maketitle
\lhead{} 

\begin{center}
\begin{minipage}{\textwidth}
\centering
\includegraphics[width=\textwidth]{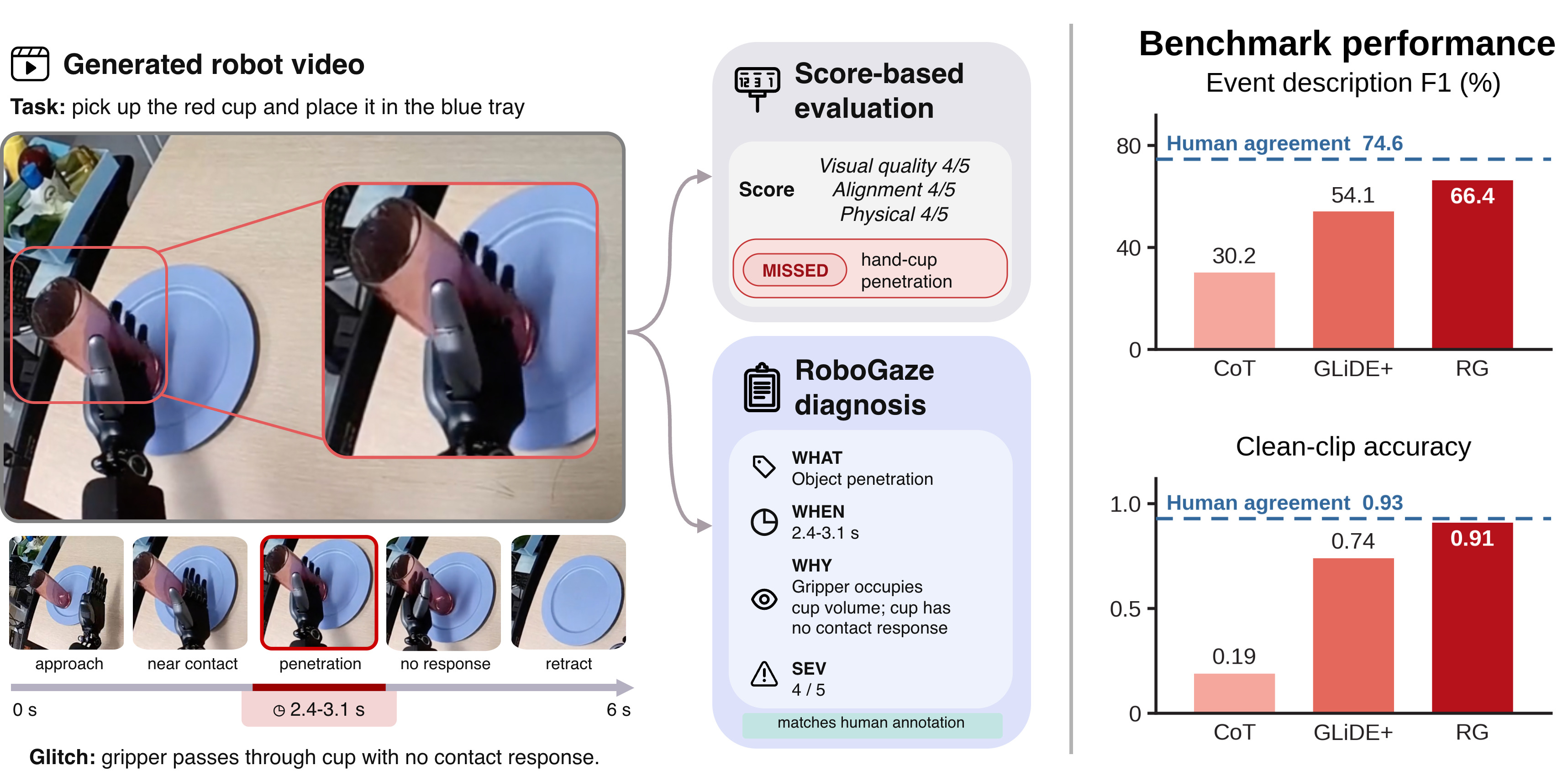}
\captionof{figure}{\textbf{From scalar judgment to grounded diagnosis.} \emph{Left:}
VideoScore2~\citep{he2025videoscore2}, a learned score-based judge, rates the clip
$4/5$ on visual quality, alignment and physical consistency, with no indication of
what failed or when. \textsc{RoboGaze} identifies the hand--cup penetration,
localizes it to $2.4$--$3.1$\,s, explains the supporting evidence and grades its
severity. \emph{Right:} against evaluators that do report events (CoT and \gplus),
on Gemini~3.1~Pro over all $914$ clips, \textsc{RoboGaze} improves description
$F_1$ and clean-clip accuracy. Dashed lines mark human agreement on the
$180$-clip gold subset; on that subset, model scores are within $0.9$ $F_1$ of
their full-benchmark values.}
\label{fig:teaser}
\end{minipage}
\end{center}

\begin{abstract}
Robot world models now generate synthetic manipulation videos, but evaluating them is hard: realistic-looking outputs often violate physical laws, temporal consistency or task logic, while conventional metrics and monolithic Vision-Language Model (VLM) judges return scores that do not say what failed. We present RoboGaze, a training-free multi-agent VLM evaluator that grounds the task and scene, routes each subtask to dimension-specific specialists and verifies their findings with a critic, producing temporally localized diagnostic reports that state what failed, when, why and how severely, under a schema of 6 dimensions and 30 failure types whose annotation reliability and coverage we measure. To benchmark it, we introduce RoboGazeBench, $914$ human-annotated clips from five generators, conditioned on simulated and real-world data and running from $4.9$ to $148.9$\,s. Across eight open-source and proprietary VLM backbones, RoboGaze outperforms every baseline; even a published agentic detector given the same schema and a slightly larger inference budget trails it by $10.1$--$12.3$ description-$F_1$ points, a margin that blinded human raters confirm. On the strongest backbone it closes $83\%$ of the gap between chain-of-thought prompting and human agreement, and its critic curbs the ``cry-wolf'' false positives of standard VLMs, lifting mean clean-clip accuracy from $17\%$ to $85\%$. The advantage is largest on held-out generators and long clips, and carries over to selecting generated demonstrations ($80\%$ vs.\ $71\%$ precision). 
\newline \textbf{Project page}: {\setlength{\fboxsep}{1.5pt}\colorbox{rgpinkbg}{\small\hypersetup{urlcolor=rgpink}\href{https://robogazeproject.github.io/RoboGazeProject/}{\texttt{robogazeproject.github.io/RoboGazeProject}}}}.
\end{abstract}

\section{Introduction}

Recent advances in generative world models (hereafter \emph{generators}) have made synthetic video a promising interface for embodied AI, enabling robot-centric rollouts for manipulation prediction, demonstration synthesis, and policy evaluation~\citep{ali2025world, hu2023gaia, bruce2024genie, yang2023unisim, hongcogvideo}. These rollouts increasingly stand in for teleoperated demonstrations and policy-evaluation episodes, so each individual clip must be correct. For robotics, however, visual realism alone is insufficient: generated videos must also be physically plausible, temporally coherent, and task-executable. We call a video \emph{physically} plausible when the robot body and the objects persist coherently and their contact is supportable (no penetration, unsupported motion or teleportation). Such failures are often brief and local, like a hand passing through a cup for a fraction of a second, so a clip can look convincing overall yet contain exactly the error that matters, a gap between perceptual quality and robotics utility.

Existing evaluation methods do not fully capture these requirements. Perceptual metrics measure low-level visual fidelity, while general video evaluators such as VBench~\citep{huang2024vbench}, DOVER~\citep{wu2023exploring}, and VideoScore~\citep{he2024videoscore} primarily assess overall video quality rather than execution correctness. Robotics benchmarks such as RBench~\citep{deng2026rethinking}, WorldSimBench~\citep{qin2024worldsimbench}, and PAI-Bench~\citep{paibench} evaluate more relevant physical or task properties, but mainly aggregate scores for model-level comparison rather than diagnose individual clips. Human evaluation~\citep{zhang2026physion} is more reliable but costly to scale.

Recent VLM-based evaluators~\citep{he2025videoscore2, mou2025gradeo, inbasekar2026worldjen} and world critics~\citep{zheng2025vbench, zhang2025morpheus, jiang2026robowm} improve semantic and physical reasoning, yet typically produce holistic judgments and can over-report failures when explicitly prompted to search for them (\S\ref{sec:main_results}). Event-level methods such as GLiDE~\citep{zheng2026open} and Spotlight~\citep{gupta2026spotlight} localize failures in time, but target game or text-to-video content, use fixed error categories, and do not condition on task instructions. Thus, robotics-specific per-video diagnosis remains underexplored: identifying \emph{what} failed, \emph{when} it failed, \emph{why} it failed, and \emph{how severe} the failure was.

To address this gap, we introduce \textbf{RoboGaze}, a training-free multi-agent VLM evaluator for robot video generation (Figure~\ref{fig:teaser}). RoboGaze is built around three key innovations: (i) \textit{task- and scene-grounded context construction} that separates what the instruction expects from what the video shows and aligns the video with its subtasks; (ii) \textit{routed, dimension-specific specialists} that localize failures within each subtask; and (iii) \textit{a critic verifier} that validates, merges and reports failures with interpretable evidence. Grounding supplies the task context that event detectors lack, routing localizes each failure in time, and the critic suppresses the false reports of a monolithic judge. All three operate over a \emph{failure schema} whose six dimensions cover the task (\emph{Task Progress}, \emph{Instruction Consistency}), the objects and scene (\emph{Object--Scene Consistency}), the robot (\emph{Robot-Body Consistency}), their interaction (\emph{Physical Plausibility}) and whether the frames can decide at all (\emph{Visual Quality}); they are refined into $30$ observable failure types, which also define the annotation of our benchmark. Unlike every evaluator in Table~\ref{tab:related_work_comparison}, RoboGaze is at once training-free, robotics-specific and temporally localized.

To support this task, we construct \textbf{RoboGazeBench}, a human-validated benchmark of $914$ generated robot-manipulation videos. Its clips come from five generators under four generation mechanisms, from robot-specific models (Cosmos-Predict2.5, DreamGen, DreamDojo) to general-purpose image-to-video models (Wan~2.2, Veo) that we held out from development. Where prior benchmarks judge a clip as a whole, every failure here is annotated from scratch under the schema above, with a temporal span, a severity score and a free-text description, so the benchmark tests whether an evaluator finds the right failure at the right time, not only whether it rates the clip as good or bad. We also measure how reliably annotators apply the schema and how much of the phenomenon it covers (\S\ref{sec:annotation}).

Across eight frozen backbones, RoboGaze outperforms every baseline, including a published agentic detector given the same schema, the same rubric and a larger inference budget (\S\ref{sec:main_results}); controls matched in information, compute and agentic design each close part of the gap, but none of them approaches its reliability on failure-free clips (\S\ref{sec:mechanism}). The advantage is largest on the two held-out generators and long clips, is confirmed by blinded human raters, and carries over to screening generated demonstrations (\S\ref{sec:robustness}).

\begin{contribbox}
In summary, we make three contributions to robot-video evaluation:

\begin{itemize}

\item \textbf{RoboGaze.}
We propose a training-free, model-agnostic multi-agent VLM evaluator that performs temporally grounded failure diagnosis and generates structured diagnostic reports instead of scalar quality scores.

\item \textbf{RoboGazeBench.}
A human-validated benchmark of $914$ generated robot-manipulation videos from five
generators (two held out from development) under four generation mechanisms, spanning $4.9$ to $148.9$ seconds: the longest
horizon of any comparable evaluation set.

\item \textbf{Evidence that structure drives the gain.}
On eight proprietary and open-source VLMs, controls matched in information, compute and agentic design, a blinded human audit and a demonstration-screening study show that RoboGaze's gains come from how the evaluation is organised and carry over to data selection.

\end{itemize}
\end{contribbox}

\section{Related Work}
\vspace{-0.1in}
\FloatBarrier
\begin{table*}[t]
\centering
\caption{Comparison of related video evaluation methods. \textsc{RoboGaze}
(last row) is the only training-free method that provides robotics-specific,
temporally localized diagnostic reports with severity, task-progress and
physical-interaction checks. Columns are defined below the table.
}
\resizebox{\textwidth}{!}{
\setlength{\tabcolsep}{2.5pt}
\renewcommand{\arraystretch}{1.12}
\begin{tabular}{lcccccccccc}
\toprule
\textbf{Method} & \textbf{Focus} & \shortstack{\textbf{Eval. clip}\\\textbf{length}} & \shortstack{\textbf{Temp.}\\\textbf{loc.}} & \textbf{Diag.} & \textbf{Sev.} & \shortstack{\textbf{Task}\\\textbf{prog.}} & \shortstack{\textbf{Phys.}\\\textbf{inter.}} & \textbf{Robot.} & \shortstack{\textbf{Train-}\\\textbf{free}} & \textbf{Output} \\
\midrule
VBench \citep{huang2024vbench} & Gen. quality & $2.0\,$s & \xmark & \xmark & \xmark & \xmark & \xmark & \xmark & \cmark & Visual score \\
VideoScore \citep{he2024videoscore} & Gen. quality & $1.0$--$3.0\,$s & \xmark & \pmark & \xmark & \xmark & \pmark & \xmark & \xmark & Multi-dim. score \\
WORLDJEN \citep{inbasekar2026worldjen} & Multi-dim. eval. & $\sim$5--$8\,$s & \xmark & \pmark & \xmark & \pmark & \cmark & \xmark & \cmark & Multi-dim. score \\
GRADEO \citep{mou2025gradeo} & Human-like eval. & $1.0$--$5.1\,$s & \xmark & \pmark & \pmark & \pmark & \pmark & \xmark & \xmark & Semantic score \\
VideoGen-Eval \citep{yang2025videogen} & Structured eval. & -- & \pmark & \pmark & \xmark & \pmark & \pmark & \xmark & \cmark & Multi-dim. score \\
PAI-Bench \citep{paibench} & Physical realism & $\sim$2--$8\,$s & \xmark & \pmark & \xmark & \pmark & \cmark & \pmark & \cmark & Physics score \\
VideoHallu \citep{li2026videohallu} & Hallucination & $5.3\,$s avg. & \xmark & \cmark & \xmark & \xmark & \cmark & \xmark & \cmark & Hallu. report \\
GlitchBench \citep{taesiri2024glitchbench} & Glitch detection & -- & \xmark & \cmark & \xmark & \xmark & \pmark & \xmark & \cmark & Glitch report \\
GLiDE \citep{zheng2026open} & Game glitch (open) & $19.1/60\,$s$^{\dagger}$ & \cmark & \cmark & \xmark & \xmark & \pmark & \xmark & \cmark & Glitch report \\
Spotlight \citep{gupta2026spotlight} & Gen. error loc. & $\sim$5--$10\,$s & \cmark & \cmark & \cmark & \pmark & \cmark & \xmark & \cmark & Error report \\
WorldSimBench \citep{qin2024worldsimbench} & World model eval. & -- & \xmark & \pmark & \xmark & \cmark & \pmark & \cmark & \xmark & Benchmark score \\
RBench \citep{deng2026rethinking} & Robot video eval. & $4$--$8\,$s$^{*}$ & \xmark & \pmark & \xmark & \cmark & \cmark & \cmark & \cmark & Robotic score \\
RoboWM-Bench  \citep{jiang2026robowm} & Robot exec.\ eval. & $5$--$25\,$s & \pmark & \pmark & \xmark & \cmark & \pmark & \cmark & \pmark & Exec. score \\
Morpheus \citep{zhang2025morpheus} & Physical reasoning & -- & \xmark & \pmark & \xmark & \xmark & \cmark & \xmark & \cmark & Physics score \\
\midrule
\rowcolor{rgaccent}
\textbf{RoboGaze} & \textbf{Robot diagnosis} & \textbf{4.9--148.9\,s} & \cmark & \cmark & \cmark & \cmark & \cmark & \cmark & \cmark & \textbf{Diagnostic report} \\
\bottomrule
\end{tabular}
}
\vspace{0.1in}
\parbox{\textwidth}{\footnotesize
$^{\dagger}$Average/maximum. $^{*}$RBench range covers configurations with stated durations; several
robotics-specific configurations report frame rate only. Among benchmarks that report duration,
\textsc{RoboGazeBench} has the longest maximum horizon ($148.9$\,s); GlitchBench is image-only. \emph{Temp.\ loc.}: localizes failures to time spans. \emph{Diag.}: names what failed, beyond a score. \emph{Sev.}: grades how serious each failure is. \emph{Task prog.}: checks whether the instructed task or subtask is achieved, beyond matching the prompt's content. \emph{Phys.\ inter.}: checks physical laws, contact or object--robot interaction. \emph{Robot.}: designed for, or evaluated on, robot-manipulation video. \cmark: yes; \xmark: no; \pmark: limited, i.e., covered partially or only as one term of a holistic score. --: not reported.}
\label{tab:related_work_comparison}
\end{table*}
\paragraph{Robot World Models and Generated Robot Video.} Large-scale video diffusion systems such as CogVideo~\citep{hongcogvideo} enable high-fidelity synthetic videos but prioritize visual quality over physical executability~\citep{wangsurvey}. Cosmos-Predict~\citep{ali2025world} unifies multi-modal world generation with better instruction following and physical consistency, while DreamGen~\citep{jang2025dreamgen} and DreamDojo~\citep{gao2026dreamdojo} extend robot-centric generation to zero-shot and dexterous tasks. Their rollouts are increasingly consumed downstream, standing in for teleoperated demonstrations and supplying predicted dynamics for policy learning, so the correctness of an individual clip matters. Persistent failure modes such as hallucinated physics and instruction-irrelevant rollouts therefore motivate evaluation beyond visual fidelity.


\paragraph{Evaluating Generated Video.}
Traditional metrics such as PSNR, SSIM~\citep{wang2004image}, LPIPS~\citep{zhang2018unreasonable}, and FVD~\citep{unterthiner2018towards} measure perceptual or distributional similarity but largely miss physical plausibility, temporal consistency, and task correctness. More structured benchmarks, including VBench~\citep{huang2024vbench, zheng2025vbench, wu2023exploring, han2025video}, introduce multi-dimensional evaluation, while learned judges such as VideoScore and VideoScore2~\citep{he2024videoscore, he2025videoscore2} rely on large-scale supervision. Recent methods instead prompt frozen VLMs for structured critique~\citep{yang2025videogen, inbasekar2026worldjen, mou2025gradeo, qi2025vcr}, and physics-oriented benchmarks such as PAI-Bench~\citep{paibench} and Morpheus~\citep{zhang2025morpheus} characterize physical failures for model-level comparison.

However, these approaches primarily score an entire video or a predefined evaluation dimension. They provide limited support for diagnosing \emph{where} a failure occurs, \emph{what} type of failure it is, and \emph{how severe} it is within an individual execution. This distinction is particularly important for robotics, where temporally grounded and physics-sensitive judgments remain challenging even for frontier multimodal models~\citep{zhang2026physion}.

\paragraph{Localized Diagnosis and Robotics-Specific Evaluation.}

Two lines approach per-video diagnosis. Generic video evaluators localize hallucinations or glitches at the event level~\citep{li2026videohallu, taesiri2024glitchbench, zheng2026open, gupta2026spotlight}, but are not instruction-aware and rely on fixed failure taxonomies whose reliability and coverage are typically unclear. Robot-video benchmarks instead assess structural consistency, action completeness, simulation alignment, or closed-loop success~\citep{deng2026rethinking, jiang2026robowm, qin2024worldsimbench}, but primarily rank generative models through aggregate scores rather than diagnose individual executions, and simulation-based methods do not directly apply to offline real-robot videos.

RoboGaze bridges these directions with a robotics-specific, instruction-aware failure schema derived from prior physical and robot-video evaluation criteria~\citep{paibench, zhang2025morpheus, deng2026rethinking, jiang2026robowm, qin2024worldsimbench}. It measures the schema's reliability and coverage and, without training or simulation, produces structured per-video diagnoses with temporal localization and severity-aware evidence.

\vspace{-0.1in}
\section{Method}
\label{sec:method}
\vspace{-0.1in}

\begin{figure*}[t]
\centering
\includegraphics[width=\textwidth]{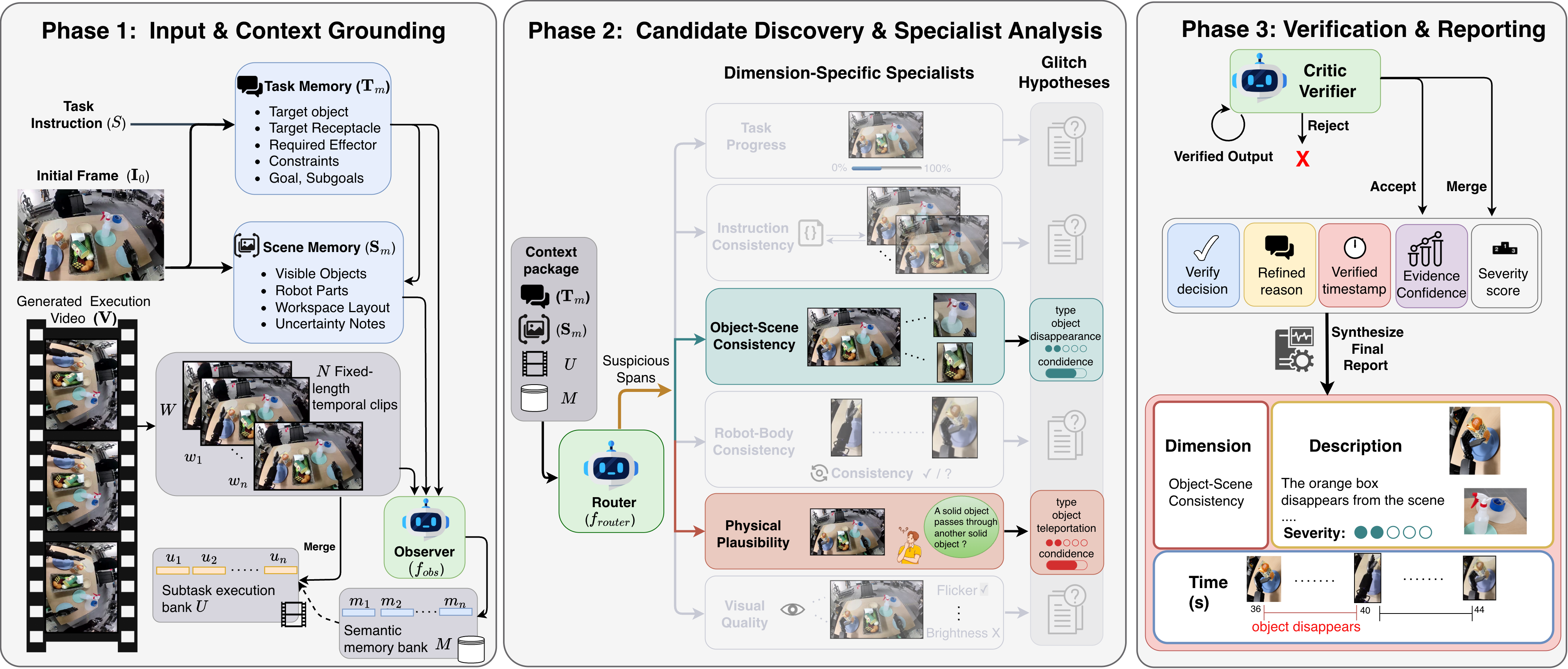}
\caption{\textbf{RoboGaze Overview}. A three-phase pipeline for video generation diagnosis: (1) extracting task and scene context memories; (2) routing suspicious temporal spans to six dimension-specific specialists to generate failure hypotheses; and (3) verifying and synthesizing hypotheses into a final structured diagnostic report.}
\label{fig:pipeline}
\end{figure*}

This section presents \textbf{RoboGaze}, a multi-agent framework for fine-grained evaluation of robot video generation. Given a task instruction, an initial scene observation, and a generated robot execution video, RoboGaze detects execution failures and produces structured diagnostic reports with failure types, supporting evidence, and likely causes. As shown in Figure~\ref{fig:pipeline} and pseudo code in appendix \ref{app:pseudocode}, RoboGaze decomposes anomaly reasoning into three phases, \textit{(1) Input and Context Grounding} (\S\ref{sec:grounding}), \textit{(2) Candidate Discovery and Specialist Analysis} (\S\ref{sec:specialists}) and \textit{(3) Verification and Reporting} (\S\ref{sec:verification}), all organised around the failure schema of \S\ref{sec:taxonomy}.

\subsection{Failure Schema}
\label{sec:taxonomy}

\paragraph{Scope.} We use \emph{failure} for any observable deviation of a
generated video from a correct, physically plausible execution of the instructed
task, including rendering conditions that make the execution visually
undecidable. This is a property of the video, not of a deployed policy: an empty
report means that no failure is visible in the sampled frames, not that the
depicted behaviour is executable on a robot. Safety, action feasibility and
downstream policy harm are outside the schema.

\begin{figure}[t]
\centering
\includegraphics[width=\textwidth]{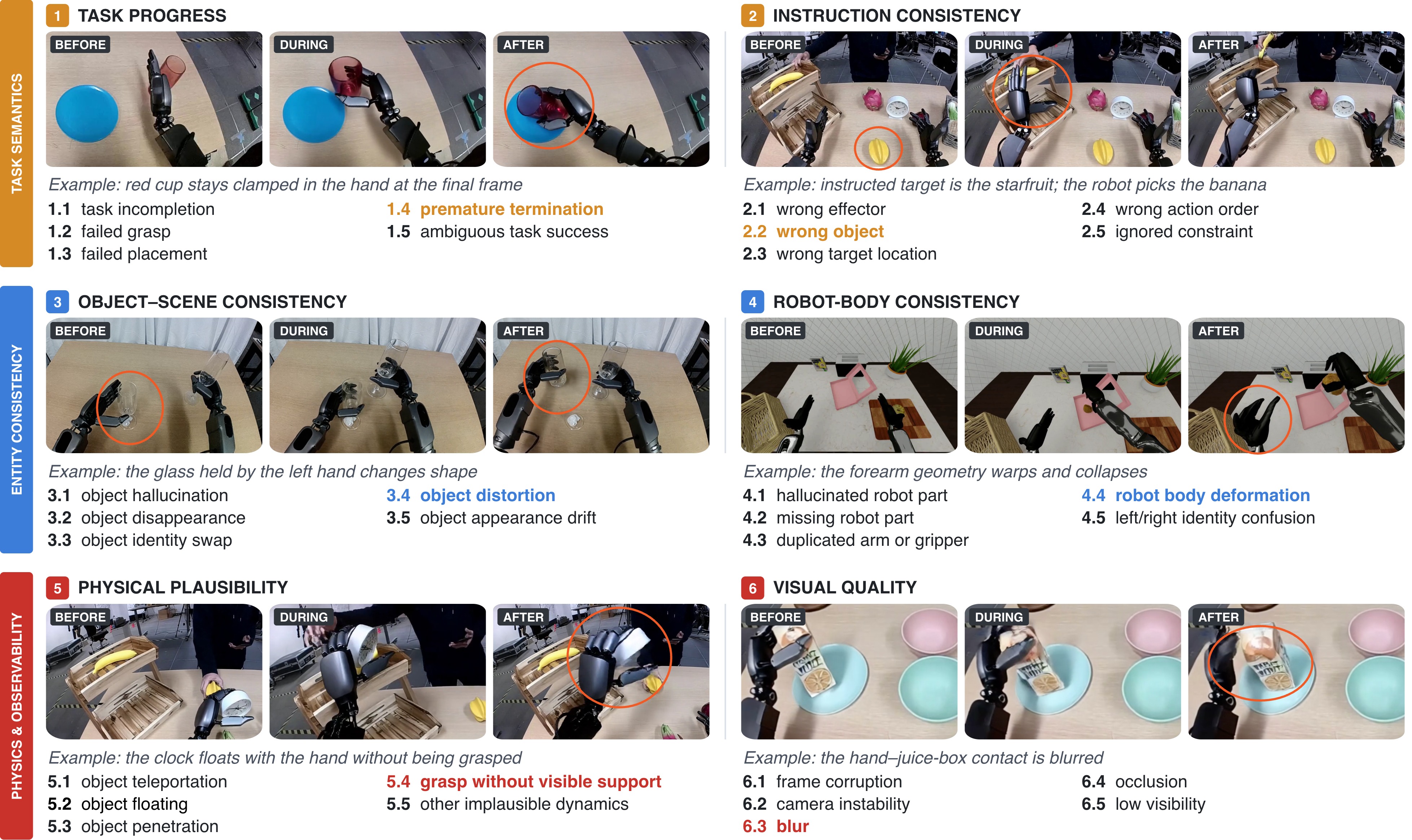}
\caption{\textbf{The \textsc{RoboGazeBench} failure schema.} Three groups (rows), each holding two dimensions, and each dimension refined into five failure types. One unmodified benchmark event per dimension is shown before, during and after the failure, and its type is highlighted. Definitions, per-type counts and the severity rubric are in Appendix~\ref{app:taxonomy}.}
\label{fig:taxonomy_schema}
\end{figure}

\paragraph{Structure.} Failures are organised in three levels (Figure~\ref{fig:taxonomy_schema}).
Three \emph{groups}, the rows of the figure, state the question being asked:
whether the video performs the right task (\emph{task semantics}), whether objects
and the robot persist coherently (\emph{entity consistency}), and whether the
result is physically supportable and visually decidable (\emph{physics and
observability}). Each group holds two \emph{dimensions}, the unit of evaluation:
one specialist handles each (\S\ref{sec:specialists}), and results are aggregated
over them because that is where annotators agree (\S\ref{sec:annotation}). Each
dimension is refined into five \emph{types}, the unit of explanation, which a
report names for every failure; three of them record that the frames cannot
decide rather than asserting a defect, so abstention is explicit. The schema was
built top-down from prior taxonomies and bottom-up from open-coded generated
clips; its construction, type definitions and exclusion rules are in
Appendix~\ref{app:taxonomy}.

\subsection{Input and Context Grounding}
\label{sec:grounding}

Given a task instruction $S$, an initial observation $I_0$, and a generated execution video $V$, RoboGaze first converts raw multimodal inputs into structured contextual memories for anomaly diagnosis, because directly prompting a VLM on long-horizon robot videos often yields weak temporal localization and ambiguous task reasoning amid large amounts of irrelevant visual context.

Specifically, RoboGaze parses $S$ and $I_0$ into a \textit{task memory} $T_m$, encoding the global task objective $o$ and an ordered list of $K$ subtask specifications, $T_m = (o, \{t_1, \dots, t_K\})$, where each $t_k$ lists the manipulated objects, target state, expected action outcome and completion criterion of subtask $k$. RoboGaze also constructs a \textit{scene memory} $S_m$ conditioned on $T_m$ and $I_0$, representing the initial workspace configuration, visible objects, robot body parts, spatial relationships, and uncertainty cues from occlusion or ambiguous object states. Together, $T_m = f_{\text{task}}(S, I_0)$ and $S_m = f_{\text{scene}}(I_0, T_m)$, where $f_{\text{task}}$ and $f_{\text{scene}}$ denote VLM-based semantic parsers, provide semantic grounding for task intent and physical grounding for scene feasibility.

To capture localized execution dynamics, RoboGaze decomposes $V$ into $N$ non-overlapping fixed-length temporal windows $\mathcal{W} = \{w_1, \dots, w_N\}$. Instead of directly localizing subtasks over the full video, RoboGaze converts localization into window-level task-progress reasoning. Each window is jointly analysed with $T_m$ and $S_m$ to produce a structured semantic observation:
\begin{equation}
\mathcal{M} = \{m_i\}_{i=1}^{N}, \quad m_i = f_{\text{obs}}(w_i, T_m, S_m),
\end{equation}
where $m_i$ summarizes robot actions, object state transitions, task progress signals, anomalies, and uncertainty cues within window $w_i$. Using these progress signals, RoboGaze groups temporally related windows into subtask-specific execution segments:
\begin{equation}
\mathcal{U} = \{u_k\}_{k=1}^{K}, \quad u_k = \mathrm{Merge}(\{w_i \mid i \in \mathcal{I}_k\}),
\end{equation}
where $\mathcal{I}_k$ denotes the windows whose progress signals assign them to subtask $t_k$, so segment $u_k$ and specification $t_k$ refer to the same subtask, and $\mathrm{Merge}(\cdot)$ combines temporally contiguous windows into a unified sub-video. Together these form the context package $\mathcal{C} = \{T_m, S_m, \mathcal{M}, \mathcal{U}\}$.

\vspace{-0.1in}
\subsection{Candidate Discovery and Specialist Analysis}
\label{sec:specialists}

Given the context package $\mathcal{C}$, RoboGaze performs targeted diagnosis with six \emph{specialists}, one per failure dimension, each reasoning only over its dimension's five types (\S\ref{sec:taxonomy}). Since only a subset of failure dimensions is typically relevant to each subtask, running all six specialists on every segment introduces noisy or weakly grounded diagnoses; RoboGaze therefore first routes each segment to the relevant specialists.

For each subtask execution segment $u_k \in \mathcal{U}$, RoboGaze retrieves the corresponding subtask specification $t_k$ from $T_m$ (\S\ref{sec:grounding}). A candidate router VLM then scores how relevant each dimension is to the segment and keeps the most relevant ones: 
\begin{equation}
\mathbf{p}_k = f_{\text{router}}(u_k, t_k) \in [0,1]^{6}, \qquad G_k^{\text{cand}} = \mathrm{TopK}(\mathbf{p}_k, 3) \subseteq \mathcal{G},
\end{equation}
where $\mathcal{G}$ denotes the six failure dimensions, so $G_k^{\text{cand}}$ is a set of at most three dimensions rather than a vector. Routing lets each specialist examine a segment closely without being asked about failures the segment cannot exhibit (e.g., a grasp segment is not checked for action order); removing the router costs $5.1$ description $F_1$ (\S\ref{sec:mechanism}).

For each candidate dimension $g \in G_k^{\text{cand}}$, the corresponding specialist $f_g$ returns a possibly empty set of hypotheses, one for each failure of its dimension it finds in the segment:
\begin{equation}
\mathcal{D}_{k,g} = f_g(u_k, t_k, S_m) = \{d_j\}, \qquad d_j = (y_j, \tau_j, r_j, e_j, \tilde{s}_j, c_j),
\end{equation}
where $y_j \in \mathcal{T}_g$ is one of the five types of dimension $g$, $\tau_j$ localizes the failure temporally, $r_j$ provides diagnostic reasoning, $e_j$ summarizes supporting evidence, $\tilde{s}_j \in \{1,\dots,5\}$ is a proposed severity, and $c_j \in [0,1]$ is the specialist's confidence. The hypotheses of all specialists, $\mathcal{D} = \bigcup_{k} \bigcup_{g \in G_k^{\text{cand}}} \mathcal{D}_{k,g}$, are passed to verification.

\vspace{-0.1in}
\subsection{Verification and Reporting}
\label{sec:verification}

The hypotheses $\mathcal{D}$ produced in Phase~2 may still be uncertain, weakly grounded, or redundant across dimensions, so a critic verifier re-examines each hypothesis $d \in \mathcal{D}$ from segment $u_k$ against the execution segment, diagnostic reasoning, temporal localization, supporting evidence, and task--scene grounding, and assigns one of three actions together with its own confidence $\hat c$:
\begin{equation}
f_{\text{verify}}(d, u_k, t_k, S_m) \;\rightarrow\; (a, \hat c), \quad a \in \{\textsc{accept},\ \textsc{reject},\ \textsc{merge}\}, \ \hat c \in [0,1].
\end{equation}

A hypothesis is \textsc{accept}ed if it is visually supported, temporally consistent, semantically compatible with the task and scene, and its confidence exceeds the fixed threshold $\tau_c=0.50$ selected on development data (Appendix~\ref{app:robustness}). It is \textsc{reject}ed when the evidence is better explained by normal execution (e.g., occlusion rather than disappearance), and \textsc{merge}d when it refers to the same physical event as another hypothesis, based on matching objects or actions, overlapping spans, and consistent evidence. Independent failures remain separate even when their spans overlap.

Each surviving hypothesis is represented as $\hat d=(y,\hat\tau,\hat r,\hat e,s)$, containing its type, verified span, refined explanation, validated evidence, and severity. The accepted and merged hypotheses form the final report $R=f_{\text{report}}(\hat{\mathcal D})$, a structured list of failure events with their dimension, type, time span, severity, description, and evidence (Figure~\ref{fig:teaser}, left; Appendix~\ref{app:record}). This verification stage filters spurious specialist predictions before reporting and is key to RoboGaze's reliability on clean clips (Section~\ref{sec:main_results}).
\vspace{-0.1in}
\section{Experimental Setup}
\label{sec:experiments}

\label{sec:benchmark}
\label{sec:annotation}
\vspace{-0.1in}

\begin{figure}[!h]
\centering
\newcommand{\keystatsminipage}[1]{\begin{minipage}[t]{0.285\linewidth}
\vspace{0pt}
\centering\rgtablefont
\setlength{\tabcolsep}{3pt}
\renewcommand{\arraystretch}{#1}
\captionof{table}{Key statistics.}
\label{tab:keystats}
\begin{tabular}{@{}lr@{}}
\toprule
Clips (failure-free) & 914 (210) \\
Annotated failure events & 1{,}847 \\
Total duration & 5.0\,h \\
Instructions & 194 \\
Generators (mechanisms) & 5 (4) \\
Embodiments & 3 \\
Duration, median (mean) & 10.4 (19.8)\,s \\
Duration, min--max & 4.9--148.9\,s \\
Clips $>40$\,s & 130 (14.2\%) \\
\bottomrule
\end{tabular}
\end{minipage}}%
\sbox{\compcapbox}{\begin{minipage}[t]{0.70\linewidth}
\vspace{0pt}
\centering
\includegraphics[width=\linewidth,trim={0 7bp 21bp 0},clip]{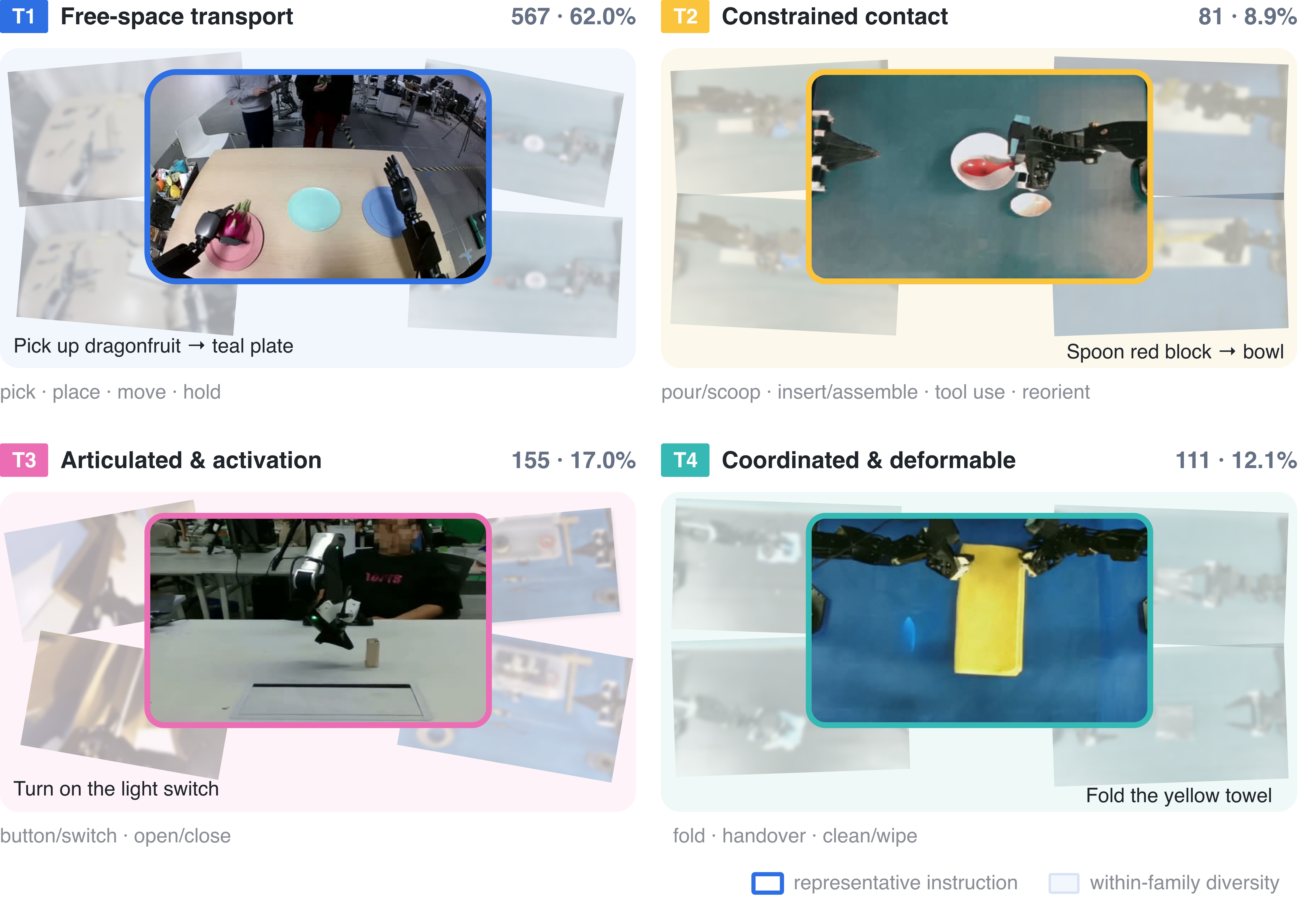}
\captionof{figure}{Task families, clip counts and example clips (Appendix~\ref{app:composition}).}
\label{fig:composition}
\end{minipage}}%
\edef\rgsavedtable{\the\value{table}}%
\sbox{\keystatsbox}{\let\label\rggobble\keystatsminipage{1.25}}%
\setcounter{table}{\rgsavedtable}%
\rgmatchstretch{\compcapbox}{\keystatsbox}{9}{1.25}%
\keystatsminipage{\rgstretch}\hfill
\usebox{\compcapbox}
\end{figure}

\paragraph{Benchmark.} \textsc{RoboGazeBench} is a set of generated robot-manipulation videos in
which every visible failure is annotated from scratch under the schema of
Section~\ref{sec:taxonomy}, with a temporal span, a severity and a free-text
description (Table~\ref{tab:keystats}, Figure~\ref{fig:composition}). Two of its
generators, Wan~2.2 and Veo ($160$ clips), were held out from all development and
differ in architecture from the Cosmos-family backbone the others share, and its
clips reach the longest horizon of any comparable evaluation set
(Table~\ref{tab:related_work_comparison}).

\paragraph{Annotation and reliability.} Each clip is labelled
independently by two robotics-trained annotators and merged by a senior annotator;
a $180$-clip \emph{gold subset}, labelled by three senior annotators, supplies the
human reference. Annotators agree on whether a clip contains a failure at Cohen's
$\kappa=.74$, and more on the dimension of a failure than on its exact type, so
results are aggregated by dimension (Appendices~\ref{app:benchmark}
and~\ref{app:taxonomy}).

\paragraph{Backbones and baselines.}\label{sec:methods}
All methods use the same eight frozen VLMs, four proprietary and four open-weight ($8$--$38$B; Table~\ref{tab:main_results}), with identical visual inputs within each backbone, enabling controlled within-backbone comparisons (Appendix~\ref{app:impl}).

To rule out simpler explanations for gains over direct prompting, we control for additional information, inference budget, and generic agentic reasoning. B1 receives the full RoboGaze context (schema, severity rubric, initial scene, and admissible normal behaviour) in a single call; B2 and B3 increase inference through self-consistency and generic propose-and-verify, respectively. B4a uses GLiDE~\citep{zheng2026open}, the closest published agentic detector, while B4b (\gplus) augments GLiDE with our schema and rubric. As the strongest and most expensive control, \gplus\ is our primary comparator in Sec.~\ref{sec:results} (Appendix~\ref{app:controls}). Learned scalar evaluators such as VideoScore2~\citep{he2025videoscore2} do not output localized events, so we evaluate them separately on candidate-clip ranking.
\paragraph{Evaluation protocol.}\label{sec:protocol}
A diagnostic report is a set of typed, time-stamped events, so we score it
event by event with the open-ended protocol of GLiDE~\citep{zheng2026open}, whose
glitch reports have the same form. A fixed LLM judge (Gemini~3.1~Pro, temperature
$0$; $r=.79$ with human ratings) rates the similarity $S_{ij}\in[0,1]$ of a predicted
and a reference description, and Hungarian assignment~\citep{kuhn1955hungarian}
pairs events one-to-one by similarity times temporal overlap $\text{IoU}_t$
(Appendix~\ref{app:metrics}). Description $F_1$ asks \emph{what} failed: precision
and recall sum $S_{ij}$ over matched pairs and divide by the number of predicted and
of reference events, so an invented event lowers precision and a missed one recall.
Temporal mIoU asks \emph{when}, as the mean overlap of matched spans, and
$F_1{\times}$IoU asks both at once, crediting each match by $S_{ij}\cdot\text{IoU}_t$.
Two judge-free metrics score what our schema adds: clean accuracy, the fraction of
the $210$ failure-free clips given an empty report, and severe-event recall (Sev-R),
the fraction of severity-$4$/$5$ reference events recovered. All are averaged over
clips; intervals are paired $95\%$ bootstrap intervals over source conditions.

\vspace{-0.1in}
\FloatBarrier
\section{Results}
\label{sec:results}
\vspace{-0.1in}

\subsection{RQ1: Does Structured Diagnosis Beat Strong Evaluators?}
\label{sec:main_results}

\begin{table}[!ht]
\centering
\footnotesize
\setlength{\tabcolsep}{1.9pt}
\renewcommand{\arraystretch}{0.93}
\caption{\textbf{Event diagnosis on \textsc{RoboGazeBench}} ($914$ clips).
RG is \textsc{RoboGaze}; \gplus\ is the schema-equipped GLiDE control (B4b).
\colorbox{rgbest}{Blue} marks the best score overall and \colorbox{rgopen}{orange}
the best among open-weight backbones;
\mgd\,\msv\,\mbz\ rank the top three. $F_1$ and $F_1{\times}$IoU are percentages; mIoU and
Clean are fractions.}
\label{tab:main_results}
\resizebox{0.9\textwidth}{!}{
\begin{tabular}{@{}llccc ccc ccc ccc@{}}
\toprule
\multirow{2}{*}{Backbone} & \multirow{2}{*}{Size}
& \multicolumn{3}{c}{Desc. $F_1 \uparrow$}
& \multicolumn{3}{c}{mIoU $\uparrow$}
& \multicolumn{3}{c}{$F_1{\times}$IoU $\uparrow$}
& \multicolumn{3}{c}{Clean $\uparrow$} \\
\cmidrule(lr){3-5}\cmidrule(lr){6-8}\cmidrule(lr){9-11}\cmidrule(lr){12-14}
& & CoT & \gplus & \textbf{RG} & CoT & \gplus & \textbf{RG} & CoT & \gplus & \textbf{RG} & CoT & \gplus & \textbf{RG} \\
\midrule
\rowcolor{rgsummary}
\multicolumn{14}{@{}l}{\textit{Proprietary backbones}} \\
\mgd\ Gemini 3.1 Pro   & -- & 30.2 & 54.1 & \cellcolor{rgbest}\textbf{66.4} & .41 & .53 & \cellcolor{rgbest}\textbf{.63} & 12.9 & 30.1 & \cellcolor{rgbest}\textbf{42.4} & .19 & .74 & \cellcolor{rgbest}\textbf{.91} \\
\msv\ GPT-5.5          & -- & 28.6 & 52.0 & \textbf{64.1} & .40 & .52 & \textbf{.62} & 12.0 & 28.4 & \textbf{40.3} & .16 & .72 & \textbf{.89} \\
\mbz\ Claude Sonnet 4.6& -- & 25.7 & 48.4 & \textbf{60.2} & .39 & .50 & \textbf{.60} & 10.5 & 25.6 & \textbf{36.7} & .18 & .70 & \textbf{.88} \\
\phantom{\mgd}\ Gemini 3.1 Flash-Lite & -- & 24.1 & 46.3 & \textbf{57.8} & .38 & .49 & \textbf{.58} & 9.6 & 23.7 & \textbf{34.1} & .14 & .67 & \textbf{.85} \\
\midrule
\rowcolor{rgsummary}
\multicolumn{14}{@{}l}{\textit{Open-weight backbones}} \\
\phantom{\mgd}\ Gemma4-31B      & 31B & 21.8 & 45.9 & \cellcolor{rgopen}\textbf{57.3} & .36 & .51 & \cellcolor{rgopen}\textbf{.62} & 8.2 & 24.8 & \cellcolor{rgopen}\textbf{36.1} & .11 & .69 & \cellcolor{rgopen}\textbf{.86} \\
\phantom{\mgd}\ Qwen3.6-35B     & 35B & 20.6 & 43.1 & \textbf{53.9} & .35 & .47 & \textbf{.57} & 7.6 & 21.6 & \textbf{31.3} & .21 & .66 & \textbf{.84} \\
\phantom{\mgd}\ InternVL3.5-38B & 38B & 19.2 & 40.2 & \textbf{50.6} & .34 & .45 & \textbf{.55} & 6.8 & 19.0 & \textbf{28.4} & .12 & .63 & \textbf{.81} \\
\phantom{\mgd}\ LLaVA-OV-2-8B   & 8B  & 17.9 & 38.0 & \textbf{48.1} & .32 & .43 & \textbf{.52} & 6.0 & 17.2 & \textbf{25.6} & .23 & .60 & \textbf{.79} \\
\midrule
\textit{Mean} & & \textit{23.5} & \textit{46.0} & \textbf{57.3} & \textit{.37} & \textit{.49} & \textbf{.59} & \textit{9.2} & \textit{23.8} & \textbf{34.4} & \textit{.17} & \textit{.68} & \textbf{.85} \\
\midrule
\rowcolor{rgrow}
\textit{Human agreement}$^{\dagger}$ & -- & \multicolumn{3}{c}{\textit{74.6}} & \multicolumn{3}{c}{\textit{.69}} & \multicolumn{3}{c}{\textit{52.2}} & \multicolumn{3}{c}{\textit{.93}} \\
\bottomrule
\end{tabular}}
\vspace{0.2em}
\\{\scriptsize $^{\dagger}$Senior annotators against each other, $180$-clip gold subset; model scores there are within $0.9$ $F_1$ of full-benchmark values (Appendix~\ref{app:gold}).}
\end{table}

\paragraph{\textsc{RoboGazeBench} is hard, and the gain is consistent across backbones.}
With chain-of-thought prompting (Table~\ref{tab:main_results}), no backbone exceeds
$30.2$ description $F_1$ or returns an empty report on more than $23\%$ of
failure-free clips, and even \gplus\ stays $20.5$ $F_1$ below human agreement on the
strongest backbone ($54.1$ against $74.6$). \textsc{RoboGaze} is best on every metric
and every backbone: averaged over the eight backbones, description $F_1$ rises from
$23.5$ with CoT and $46.0$ with \gplus\ to $57.3$, a paired gain of $+11.3$
$[10.2, 12.4]$ that stays within $+10.1$ to $+12.3$ although the backbones span
$18.3$ points. It is largest on the strongest backbone, so the structure adds to what a stronger
model knows rather than compensating for a weaker one.

\paragraph{The widest separation is on clean clips; the residual gap sits where annotators disagree.}
\label{sec:failure}
With no judge involved, CoT returns an empty report on $17\%$ of failure-free clips,
\gplus\ on $68\%$ and \textsc{RoboGaze} on $85\%$ ($91\%$ on Gemini~3.1~Pro). On the
gold subset, \textsc{RoboGaze} on Gemini~3.1~Pro recovers $83\%$ of the distance
between CoT and human agreement in description $F_1$, against $55\%$ for \gplus, and
$99\%$ of it in clean accuracy (Appendix~\ref{app:gold}). The residual concentrates in
Physical Plausibility and Visual Quality and tracks inter-annotator agreement across the six dimensions
($\rho=-.94$; Appendix~\ref{app:ablations}): part of the ceiling reflects what sampled
RGB frames can settle, not the evaluator (error atlas in Appendix~\ref{app:atlas}).

\subsection{RQ2: Where Does the Gain Come From?}
\label{sec:mechanism}

\begin{figure}[t]
\centering
\includegraphics[width=0.9\linewidth]{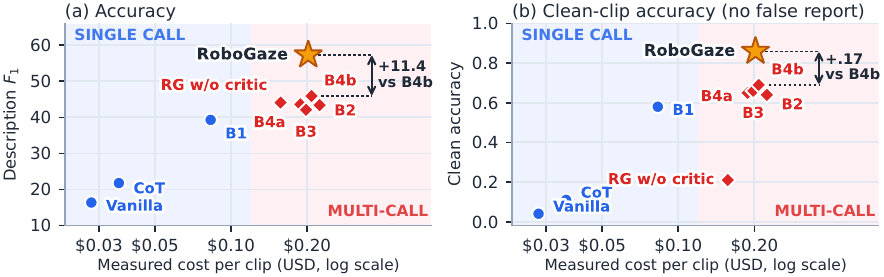}
\caption{\textbf{Structure, not information, compute or agentic design,
separates the methods} (Gemma4-31B, all $914$ clips; controls in \S\ref{sec:methods}; full values in Appendix~\ref{app:controls}).}
\label{fig:controls}
\end{figure}
\vspace{-0.05in}
\paragraph{Each control closes part of the gap, but none matches it on clean clips.}
On Gemma4-31B (Figure~\ref{fig:controls}; same ordering on Gemini~3.1~Pro,
Appendix~\ref{app:controls}), a single call given everything \textsc{RoboGaze} is
told (B1) closes $49\%$ of the $F_1$ distance from CoT to \textsc{RoboGaze}, but its clean
accuracy stays at $.58$ against $.86$: the schema tells the model what counts as a
failure, not whether the frames show one. Spending more on the same evidence adds
little: nine self-consistent samples (B2) gain $4.1$ $F_1$ over B1, and a
propose-and-verify loop (B3) does no better. GLiDE given our schema and rubric (B4b)
is the strongest control, closing $68\%$, yet at comparable cost \textsc{RoboGaze} still
leads it by $11.4$ $F_1$ and $.17$ clean accuracy.

\begin{wraptable}{r}{0.43\linewidth}
\vspace{-12pt}
\centering\footnotesize
\setlength{\tabcolsep}{3.5pt}
\caption{Component ablation, Gemma4-31B; all eight configurations in Appendix~\ref{app:ablations}.}
\label{tab:ablation}
\begin{tabular}{@{}lrrr@{}}
\toprule
Configuration & $F_1$ & Sev-R & Clean \\
\midrule
\rowcolor{rgaccent}
\textbf{\textsc{RoboGaze}} & \textbf{57.3} & .91 & \textbf{.86} \\
w/o task \& scene mem. & 50.5 & .86 & .79 \\
w/o candidate router & 52.2 & .87 & .82 \\
\rowcolor{rgdrop}
w/o critic verifier & 44.1 & \textbf{.94} & .21 \\
\bottomrule
\end{tabular}
\vspace{-10pt}
\end{wraptable}
\paragraph{Grounding and routing find events; the critic removes false ones.}
Removing task and scene memory or the candidate router costs $6.8$ and $5.1$ $F_1$ and
$.05$ and $.04$ severe-event recall (Table~\ref{tab:ablation}). Removing the critic
costs $13.2$ $F_1$ and collapses clean accuracy from $.86$ to $.21$, yet is the only
removal that \emph{raises} severe-event recall ($+.03$); the ordering holds over
thresholds $.30$--$.70$ (Appendix~\ref{app:robustness}).

\vspace{-0.05in}
\subsection{RQ3: Does It Hold Beyond the Development Conditions?}
\label{sec:robustness}
\label{sec:horizon}
\label{sec:validity}
\begin{figure}[!ht]
\centering
\includegraphics[width=0.9\linewidth]{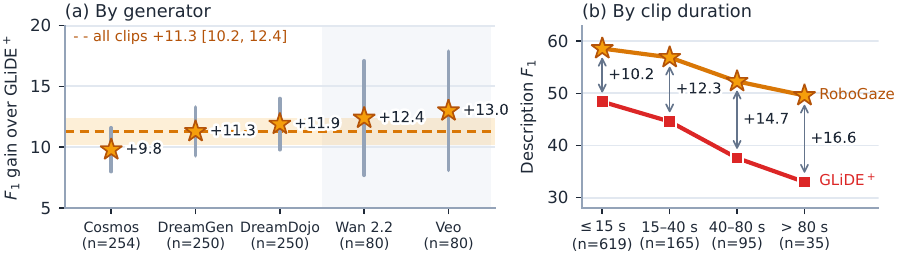}
\caption{\textbf{The margin is largest where the benchmark is hardest}
(pooled over eight backbones). (a) Paired $F_1$ gain over \gplus\ by generator, $95\%$ intervals; shaded: held out. (b) $F_1$ by clip duration (Appendix~\ref{app:slices}).}
\label{fig:robustness}
\vspace{-0.05in}
\end{figure}

\vspace{-0.05in}
\paragraph{The margin grows on held-out generators and long clips.} Both methods
score lower on Wan~2.2 and Veo, held out when prompts and thresholds were chosen, but
\gplus\ loses more: the gain is $+12.4$ and $+13.0$ $F_1$ against $+11.3$ overall
(Figure~\ref{fig:robustness}a). It also rises monotonically with clip length, from
$+10.2$ below fifteen seconds to $+16.6$ above eighty (Figure~\ref{fig:robustness}b),
and all eighteen slice intervals exclude zero (Appendix~\ref{app:slices}). On long
clips sixteen uniform frames skip $68\%$ of reference events, which \textsc{RoboGaze}
reads window by window; more frames barely help \gplus\ (Appendix~\ref{app:robustness}).

\paragraph{The scores do not depend on the language judge.} Three blind senior
annotators re-scoring \textsc{RoboGaze} and \gplus\ (Gemma4-31B) on the gold subset give
$71.6$ against $61.4$ event $F_1$, a margin of $+10.2$ close to the automatic $+11.4$ on
the same clips (Appendix~\ref{app:humanvalid}). Re-scoring every cell with GPT-5.5
preserves the ordering ($\rho=.95$, largest change $1.9$ points;
Appendix~\ref{app:robustness}).

\paragraph{The reports also select better demonstrations.}
\label{sec:utility}
Ranking six renderings of each of $40$
conditions outside the benchmark, with usability labelled by blind annotators,
\textsc{RoboGaze} keeps usable clips in its top half at $.80$ precision, against $.71$
for \gplus, $.65$ for VideoScore2~\citep{he2025videoscore2} and $.62$ for RBench, and
halves surviving severe failures ($.05$ against $.11$; Appendix~\ref{app:screening}).

\vspace{-0.1in}
\section{Conclusion and Limitations}
\label{sec:conclusion}
\vspace{-0.1in}

We presented \textsc{RoboGaze}, a training-free evaluator that turns frozen VLM judges
into temporally localized robot-video diagnosticians, and \textsc{RoboGazeBench}, the
longest-horizon benchmark of its kind. Evaluator \emph{structure}, not information,
compute or agentic design, drives a consistent gain that grows on held-out
generators and long clips.

\paragraph{Limitations.}\label{sec:limitations}
\textsc{RoboGaze} reports \emph{visible} failures: an empty report is not
evidence of physical executability and does not replace human oversight before
real-robot use. The residual gap to humans lies in \emph{Physical Plausibility} and
\emph{Visual Quality}, where annotators also disagree most, suggesting that 3D or
pixel-level signals are needed beyond sampled RGB frames. All clips are single-view, whereas recent robot
world models also render multi-camera rollouts; judging cross-view consistency,
which remains hard for current VLMs, is left to future work. Finally, the precision-oriented critic can suppress rare
transient failures, and the multi-call design costs more than a single judge.

\section*{AI Use Statement}
Generative video models were used to create the synthetic videos that constitute
\textsc{RoboGazeBench}, and vision--language models were used as the experimental
systems and evaluators studied in this work, as described in
Section~\ref{sec:experiments} and
Appendices~\ref{app:benchmark} and~\ref{app:implmetrics}. Generative AI tools were also
used to assist with literature review and to polish the wording of the
manuscript. The authors checked cited works against their primary sources,
verified numerical claims against the experiment records, and reviewed all
AI-assisted text. The authors take responsibility for the complete content of
this work.

\section*{Ethics Statement}
This work studies generated robot videos and does not evaluate people or deploy
an autonomous system in the physical world. Seven annotators with relevant
technical backgrounds labelled the videos after training and calibration; they
were paid above local hourly research-assistant rates. The activity was reviewed
internally as non-human-subject work. Annotators saw only generated videos and
task instructions and were blinded to generator and evaluator identity. Released
annotation records use pseudonymous annotator identifiers and exclude directly
identifying information. Source datasets are redistributed only under their
original licences. The benchmark may contain generator- and task-dependent
biases and does not certify physical executability or safety: false negatives
remain possible, and its predictions should not replace human oversight in a
real-robot setting. Further details on annotation, release contents, and intended
use appear in Appendices~\ref{app:annotation} and~\ref{app:atlas}.

\section*{Reproducibility Statement}
Section~\ref{sec:experiments} specifies benchmark construction,
annotation, baselines, metrics, and statistical testing. The appendix records
the generation and filtering procedure (Appendix~\ref{app:benchmark}), exact
model checkpoints and API access dates, frame budgets, decoding settings,
retry policy, prompt versions and hashes, and compute configuration
(Appendix~\ref{app:impl}), as well as metric definitions and robustness analyses
(Appendices~\ref{app:metrics} and~\ref{app:robustness}). The release
contains the benchmark manifest, prompts, configurations, model outputs,
annotations, scoring code, and figure-generation scripts needed to reproduce the
reported results; it will be made public upon acceptance.

\FloatBarrier


\bibliography{ref}

@article{ali2025world,
  title={World simulation with video foundation models for Physical {AI}},
  author={Ali, Arslan and Bai, Junjie and Bala, Maciej and Balaji, Yogesh and Blakeman, Aaron and Cai, Tiffany and Cao, Jiaxin and Cao, Tianshi and Cha, Elizabeth and Chao, Yu-Wei and others},
  journal={arXiv preprint arXiv:2511.00062},
  year={2025}
}

@article{hu2023gaia,
  title={{GAIA-1}: A generative world model for autonomous driving},
  author={Hu, Anthony and Russell, Lloyd and Yeo, Hudson and Murez, Zak and Fedoseev, George and Kendall, Alex and Shotton, Jamie and Corrado, Gianluca},
  journal={arXiv preprint arXiv:2309.17080},
  year={2023}
}

@inproceedings{bruce2024genie,
  title={{Genie}: Generative interactive environments},
  author={Bruce, Jake and Dennis, Michael D and Edwards, Ashley and Parker-Holder, Jack and Shi, Yuge and Hughes, Edward and Lai, Matthew and Mavalankar, Aditi and Steigerwald, Richie and Apps, Chris and others},
  booktitle={Forty-first International Conference on Machine Learning},
  year={2024}
}

@inproceedings{yang2023unisim,
  title={{UniSim}: A neural closed-loop sensor simulator},
  author={Yang, Ze and Chen, Yun and Wang, Jingkang and Manivasagam, Sivabalan and Ma, Wei-Chiu and Yang, Anqi Joyce and Urtasun, Raquel},
  booktitle={Proceedings of the IEEE/CVF Conference on Computer Vision and Pattern Recognition},
  pages={1389--1399},
  year={2023}
}

@inproceedings{hongcogvideo,
  title={{CogVideo}: Large-scale Pretraining for Text-to-Video Generation via Transformers},
  author={Hong, Wenyi and Ding, Ming and Zheng, Wendi and Liu, Xinghan and Tang, Jie},
  booktitle={The Eleventh International Conference on Learning Representations},
  year={2023}
}

@inproceedings{huang2024vbench,
  title={{VBench}: Comprehensive benchmark suite for video generative models},
  author={Huang, Ziqi and He, Yinan and Yu, Jiashuo and Zhang, Fan and Si, Chenyang and Jiang, Yuming and Zhang, Yuanhan and Wu, Tianxing and Jin, Qingyang and Chanpaisit, Nattapol and others},
  booktitle={Proceedings of the IEEE/CVF Conference on Computer Vision and Pattern Recognition},
  pages={21807--21818},
  year={2024}
}

@article{zheng2025vbench,
  title={{VBench-2.0}: Advancing video generation benchmark suite for intrinsic faithfulness},
  author={Zheng, Dian and Huang, Ziqi and Liu, Hongbo and Zou, Kai and He, Yinan and Zhang, Fan and Gu, Lulu and Zhang, Yuanhan and He, Jingwen and Zheng, Wei-Shi and others},
  journal={arXiv preprint arXiv:2503.21755},
  year={2025}
}

@inproceedings{wu2023exploring,
  title={Exploring video quality assessment on user generated contents from aesthetic and technical perspectives},
  author={Wu, Haoning and Zhang, Erli and Liao, Liang and Chen, Chaofeng and Hou, Jingwen and Wang, Annan and Sun, Wenxiu and Yan, Qiong and Lin, Weisi},
  booktitle={Proceedings of the IEEE/CVF international conference on computer vision},
  pages={20144--20154},
  year={2023}
}

@inproceedings{he2024videoscore,
  title={{VideoScore}: Building automatic metrics to simulate fine-grained human feedback for video generation},
  author={He, Xuan and Jiang, Dongfu and Zhang, Ge and Ku, Max and Soni, Achint and Siu, Sherman and Chen, Haonan and Chandra, Abhranil and Jiang, Ziyan and Arulraj, Aaran and others},
  booktitle={Proceedings of the 2024 Conference on Empirical Methods in Natural Language Processing},
  pages={2105--2123},
  year={2024}
}

@article{he2025videoscore2,
  title={{VideoScore2}: Think before you score in generative video evaluation},
  author={He, Xuan and Jiang, Dongfu and Nie, Ping and Liu, Minghao and Jiang, Zhengxuan and Su, Mingyi and Ma, Wentao and Lin, Junru and Ye, Chun and Lu, Yi and others},
  journal={arXiv preprint arXiv:2509.22799},
  year={2025}
}

@article{zhang2026physion,
  title={{Physion-Eval}: Evaluating Physical Realism in Generated Video via Human Reasoning},
  author={Zhang, Qin and Jing, Peiyu and Yu, Hong-Xing and Ding, Fangqiang and Nie, Fan and Wang, Weimin and Du, Yilun and Zou, James and Wu, Jiajun and Shuai, Bing},
  journal={arXiv preprint arXiv:2603.19607},
  year={2026}
}

@inproceedings{mou2025gradeo,
  title={{GRADEO}: Towards Human-Like Evaluation for Text-to-Video Generation via Multi-Step Reasoning},
  author={Mou, Zhun and Xia, Bin and Huang, Zhengchao and Yang, Wenming and Jia, Jiaya},
  booktitle={International Conference on Machine Learning},
  pages={44971--44996},
  year={2025},
  organization={PMLR}
}

@article{inbasekar2026worldjen,
  title={{WorldJen}: An End-to-End Multi-Dimensional Benchmark for Generative Video Models},
  author={Inbasekar, Karthik and Rom, Guy and Shlomovits, Omer},
  journal={arXiv preprint arXiv:2605.03475},
  year={2026}
}

@article{zhang2025morpheus,
  title={{Morpheus}: Benchmarking physical reasoning of video generative models with real physical experiments},
  author={Zhang, Chenyu and Cherniavskii, Daniil and Tragoudaras, Antonios and Vozikis, Antonios and Nijdam, Thijmen and Prinzhorn, Derck WE and Bodracska, Mark and Sebe, Nicu and Zadaianchuk, Andrii and Gavves, Efstratios},
  journal={arXiv preprint arXiv:2504.02918},
  year={2025}
}

@article{jiang2026robowm,
  title={{RoboWM-Bench}: A Benchmark for Evaluating World Models in Robotic Manipulation},
  author={Jiang, Feng and Chen, Yang and Xu, Kyle and Liu, Yuchen and Wang, Haifeng and Shen, Zhenhao and Lu, Jasper and Huang, Shengze and Wang, Yuanfei and Xie, Chen and others},
  journal={arXiv preprint arXiv:2604.19092},
  year={2026}
}

@article{wang2004image,
  title={Image quality assessment: from error visibility to structural similarity},
  author={Wang, Zhou and Bovik, Alan C and Sheikh, Hamid R and Simoncelli, Eero P},
  journal={IEEE transactions on image processing},
  volume={13},
  number={4},
  pages={600--612},
  year={2004},
  publisher={IEEE}
}

@inproceedings{zhang2018unreasonable,
  title={The unreasonable effectiveness of deep features as a perceptual metric},
  author={Zhang, Richard and Isola, Phillip and Efros, Alexei A and Shechtman, Eli and Wang, Oliver},
  booktitle={Proceedings of the IEEE conference on computer vision and pattern recognition},
  pages={586--595},
  year={2018}
}

@article{unterthiner2018towards,
  title={Towards accurate generative models of video: A new metric \& challenges},
  author={Unterthiner, Thomas and Van Steenkiste, Sjoerd and Kurach, Karol and Marinier, Raphael and Michalski, Marcin and Gelly, Sylvain},
  journal={arXiv preprint arXiv:1812.01717},
  year={2018}
}

@inproceedings{jang2025dreamgen,
  title={{DreamGen}: Unlocking Generalization in Robot Learning through Video World Models},
  author={Jang, Joel and Ye, Seonghyeon and Lin, Zongyu and Xiang, Jiannan and Bjorck, Johan and Fang, Yu and Hu, Fengyuan and Huang, Spencer and Kundalia, Kaushil and Lin, Yen-Chen and others},
  booktitle={Conference on Robot Learning},
  pages={5170--5194},
  year={2025},
  organization={PMLR}
}

@article{gao2026dreamdojo,
  title={{DreamDojo}: A Generalist Robot World Model from Large-Scale Human Videos},
  author={Gao, Shenyuan and Liang, William and Zheng, Kaiyuan and Malik, Ayaan and Ye, Seonghyeon and Yu, Sihyun and Tseng, Wei-Cheng and Dong, Yuzhu and Mo, Kaichun and Lin, Chen-Hsuan and others},
  journal={arXiv preprint arXiv:2602.06949},
  year={2026}
}

@article{wangsurvey,
  title={Survey of Video Diffusion Models: Foundations, Implementations, and Applications},
  author={Wang, Yimu and Liu, Xuye and Pang, Wei and Ma, Li and Yuan, Shuai and Debevec, Paul and Yu, Ning},
  journal={Transactions on Machine Learning Research},
  year={2025}
}

@inproceedings{han2025video,
  title={{Video-Bench}: Human-aligned video generation benchmark},
  author={Han, Hui and Li, Siyuan and Chen, Jiaqi and Yuan, Yiwen and Wu, Yuling and Deng, Yufan and Leong, Chak Tou and Du, Hanwen and Fu, Junchen and Li, Youhua and others},
  booktitle={Proceedings of the Computer Vision and Pattern Recognition Conference},
  pages={18858--18868},
  year={2025}
}

@article{yang2025videogen,
  title={{VideoGen-Eval}: Agent-based system for video generation evaluation},
  author={Yang, Yuhang and Fan, Ke and Sun, Shangkun and Li, Hongxiang and Zeng, Ailing and Han, FeiLin and Zhai, Wei and Liu, Wei and Cao, Yang and Zha, Zheng-Jun},
  journal={arXiv preprint arXiv:2503.23452},
  year={2025}
}

@article{qi2025vcr,
  title={{VCR-Bench}: A comprehensive evaluation framework for video chain-of-thought reasoning},
  author={Qi, Yukun and Zhao, Yiming and Zeng, Yu and Bao, Xikun and Huang, Wenxuan and Chen, Lin and Chen, Zehui and Zhao, Jie and Qi, Zhongang and Zhao, Feng},
  journal={arXiv preprint arXiv:2504.07956},
  year={2025}
}

@article{li2026videohallu,
  title={{VideoHallu}: Evaluating and mitigating multi-modal hallucinations on synthetic video understanding},
  author={Li, Zongxia and Wu, Xiyang and Shi, Guangyao and Qin, Yubin and Du, Hongyang and Zhou, Tianyi and Manocha, Dinesh and Boyd-Graber, Jordan},
  journal={Advances in Neural Information Processing Systems},
  volume={38},
  pages={76046--76078},
  year={2025}
}

@article{deng2026rethinking,
  title={Rethinking Video Generation Model for the Embodied World},
  author={Deng, Yufan and Pan, Zilin and Zhang, Hongyu and Li, Xiaojie and Hu, Ruoqing and Ding, Yufei and Zou, Yiming and Zeng, Yan and Zhou, Daquan},
  journal={arXiv preprint arXiv:2601.15282},
  year={2026}
}

@inproceedings{taesiri2024glitchbench,
  title={{GlitchBench}: Can large multimodal models detect video game glitches?},
  author={Taesiri, Mohammad Reza and Feng, Tianjun and Bezemer, Cor-Paul and Nguyen, Anh},
  booktitle={Proceedings of the IEEE/CVF Conference on Computer Vision and Pattern Recognition},
  pages={22444--22455},
  year={2024}
}

@article{zheng2026open,
  title={Open-Ended Video Game Glitch Detection with Agentic Reasoning and Temporal Grounding},
  author={Zheng, Muyang and Zhou, Tong and Wu, Geyang and Lin, Zihao and Wang, Haibo and Huang, Lifu},
  journal={arXiv preprint arXiv:2604.07818},
  year={2026}
}

@misc{paibench,
      title={{PAI-Bench}: A Comprehensive Benchmark For Physical {AI}}, 
      author={Fengzhe Zhou and Jiannan Huang and Jialuo Li and Deva Ramanan and Humphrey Shi},
      year={2025},
      eprint={2512.01989},
      archivePrefix={arXiv},
      primaryClass={cs.CV},
      url={https://arxiv.org/abs/2512.01989}, 
}

@article{qin2024worldsimbench,
  title={{WorldSimBench}: Towards Video Generation Models as World Simulators},
  author={Qin, Yiran and Shi, Zhelun and Yu, Jiwen and Wang, Xijun and Zhou, Enshen and Li, Lijun and Yin, Zhenfei and Liu, Xihui and Sheng, Lu and Shao, Jing and others},
  journal={arXiv preprint arXiv:2410.18072},
  year={2024}
}

@article{kuhn1955hungarian,
  title={The {H}ungarian method for the assignment problem},
  author={Kuhn, Harold W},
  journal={Naval research logistics quarterly},
  volume={2},
  number={1-2},
  pages={83--97},
  year={1955},
  publisher={Wiley Online Library}
}

@article{gupta2026spotlight,
  title={{Spotlight}: Identifying and Localizing Video Generation Errors Using {VLMs}},
  author={Chinchure, Aditya and Ravi, Sahithya and Shukla, Pushkar and Shwartz, Vered and Sigal, Leonid},
  journal={arXiv preprint arXiv:2511.18102},
  year={2025}
}

@article{nickerson2013taxonomy,
  title={A method for taxonomy development and its application in information systems},
  author={Nickerson, Robert C. and Varshney, Upkar and Muntermann, Jan},
  journal={European Journal of Information Systems},
  volume={22},
  number={3},
  pages={336--359},
  year={2013}
}

@article{cemri2025mast,
  title={Why Do Multi-Agent {LLM} Systems Fail?},
  author={Cemri, Mert and Pan, Melissa Z. and Yang, Shuyi and others},
  journal={arXiv preprint arXiv:2503.13657},
  year={2025}
}
\bibliographystyle{iclr2027_conference}

\clearpage
\appendix
\begin{center}
{\Large\bfseries Appendix}
\end{center}

\noindent This supplement provides additional details on the RoboGaze pseudocode, failure schema,
benchmark construction and annotation, implementation and metrics, complete and
robustness results, and further qualitative examples for RoboGaze.

\vspace{1.5em}

\begingroup
\makeatletter
\newcommand{\appendixroadmapline}[2]{%
  \vspace{0.6em}%
  \@dottedtocline{1}{0em}{2.2em}{\textbf{\textcolor{rglink}{(\ref*{#1})}}\hspace{0.4em}\hyperref[#1]{\textbf{#2}}}{\textbf{\pageref{#1}}}}
\newcommand{\appendixroadmapsubline}[2]{%
  \@dottedtocline{2}{2.4em}{3.2em}{\textcolor{rglink}{(\ref*{#1})}\hspace{0.4em}\hyperref[#1]{#2}}{\pageref{#1}}}
\appendixroadmapline{app:pseudocode}{RoboGaze Pseudocode}
\appendixroadmapline{app:taxonomy}{Failure Schema}
\appendixroadmapsubline{app:schema_construction}{Scope, Construction and Status}
\appendixroadmapsubline{app:codebook}{Type Definitions and Exclusion Rules}
\appendixroadmapsubline{app:severity}{Severity, Cascading Events and Primary Dimension}
\appendixroadmapsubline{app:type_counts}{Per-Type Event Counts}
\appendixroadmapsubline{app:residual}{Events the Schema Did Not Cover}
\appendixroadmapsubline{app:coverage}{Coverage Check on Held-Out Generators}
\appendixroadmapline{app:benchmark}{Benchmark Construction and Annotation}
\appendixroadmapsubline{app:construction_overview}{Construction Overview}
\appendixroadmapsubline{app:composition}{Composition}
\appendixroadmapsubline{app:gensettings}{Generation Settings}
\appendixroadmapsubline{app:sampling}{Sampling and Matched Design}
\appendixroadmapsubline{app:filtering}{Filtering}
\appendixroadmapsubline{app:record}{Released Record Format}
\appendixroadmapsubline{app:annotation}{Annotation Protocol and Interface}
\appendixroadmapsubline{app:effort}{Annotation Effort}
\appendixroadmapsubline{app:contamination}{Contamination}
\appendixroadmapsubline{app:agreement}{Agreement}
\appendixroadmapsubline{app:agreement_levels}{Agreement at Three Granularities}
\appendixroadmapsubline{app:humanvalid}{Blinded Human Audit}
\appendixroadmapline{app:implmetrics}{Implementation and Metrics}
\appendixroadmapsubline{app:impl}{Implementation}
\appendixroadmapsubline{app:metrics}{Metrics}
\appendixroadmapline{app:results}{Complete Results and Robustness}
\appendixroadmapsubline{app:controls}{Baselines and Measured Cost}
\appendixroadmapsubline{app:slices}{Results by Slice}
\appendixroadmapsubline{app:ablations}{Complete Ablations, Per-Dimension Scores, and Judge-Free Metrics}
\appendixroadmapsubline{app:gold}{Gold Subset}
\appendixroadmapsubline{app:screening}{Offline Candidate Screening}
\appendixroadmapsubline{app:robustness}{Robustness Checks}
\appendixroadmapline{app:atlas}{Qualitative Atlas, Release, and Ethics}

\makeatother
\endgroup

\section{RoboGaze Pseudocode}
\label{app:pseudocode}

This section presents the pseudocode for the three phases of
RoboGaze: input and context grounding
(Algorithm~\ref{alg:grounding}), candidate discovery and
specialist analysis (Algorithm~\ref{alg:specialist}), and
verification and reporting (Algorithm~\ref{alg:verification}).
The phases communicate through a context package
$\mathcal{C}$ and a set of failure hypotheses $\mathcal{D}$,
producing a structured diagnostic report $R$.
All VLM components use the same frozen backbone without
task-specific training.

\begin{algorithm}[t]
\caption{Input and Context Grounding}
\label{alg:grounding}
\begin{algorithmic}[1]
\Require Task instruction $S$, initial observation $I_0$,
    video $V$
\Ensure Context package $\mathcal{C}=\{T_m,S_m,M,U\}$

\State $T_m \gets f_{\mathrm{task}}(S,I_0)$
\State $S_m \gets f_{\mathrm{scene}}(I_0,T_m)$
\State $W \gets \Call{SplitWindows}{V,\Delta=4\,\mathrm{s}}$
    \Comment{Non-overlapping windows}
\State $M \gets \emptyset$; $U \gets \emptyset$

\For{each $w_i \in W$}
    \State $m_i \gets f_{\mathrm{obs}}(w_i,T_m,S_m)$
    \State $M \gets M \cup \{m_i\}$
\EndFor

\For{each subtask $t_k$ in $T_m$}
    \State $I_k \gets \Call{AssignWindows}{M,t_k}$
        \Comment{Use task-progress signals}
    \State $u_k \gets
        \Call{MergeContiguous}{\{w_i \mid i \in I_k\}}$
    \State $U \gets U \cup \{u_k\}$
\EndFor

\State \Return $\mathcal{C}=\{T_m,S_m,M,U\}$
\end{algorithmic}
\end{algorithm}

\begin{algorithm}[t]
\caption{Candidate Discovery and Specialist Analysis}
\label{alg:specialist}
\begin{algorithmic}[1]
\Require Context $\mathcal{C}=\{T_m,S_m,M,U\}$ from
    Algorithm~\ref{alg:grounding};
    failure schema $\mathcal{G}$
\Ensure Failure hypotheses $\mathcal{D}$ with source contexts

\State $\mathcal{D} \gets \emptyset$

\For{each subtask segment $u_k \in U$}
    \State $t_k \gets \Call{RetrieveSubtask}{T_m,k}$
    \State $\mathbf{p}_k \gets f_{\mathrm{router}}(u_k,t_k)$
    \State $\mathcal{G}^{\mathrm{cand}}_k \gets
        \Call{TopKDimensions}{\mathbf{p}_k,3}$
        \Comment{At most three dimensions}

    \For{each $g \in \mathcal{G}^{\mathrm{cand}}_k$}
        \State $\mathcal{D}_{k,g} \gets f_g(u_k,t_k,S_m)$
            \Comment{Possibly empty}
        \State $\mathcal{D} \gets \mathcal{D}
            \cup \{(d,k,g) \mid d \in \mathcal{D}_{k,g}\}$
            \Comment{Retain provenance}
    \EndFor
\EndFor

\State \Return $\mathcal{D}$
\end{algorithmic}
\end{algorithm}

\begin{algorithm}[t]
\caption{Verification and Reporting}
\label{alg:verification}
\begin{algorithmic}[1]
\Require Hypotheses $\mathcal{D}$ from
    Algorithm~\ref{alg:specialist};
    context $\mathcal{C}$ from Algorithm~\ref{alg:grounding};
    thresholds $\tau_c=0.50$, $\tau_{\mathrm{IoU}}=0.60$
\Ensure Structured diagnostic report $R$

\State $\mathcal{A} \gets \emptyset$;
    $\mathcal{Q} \gets \emptyset$

\For{each $(d,k,g) \in \mathcal{D}$}
    \State $(u_k,t_k,S_m) \gets
        \Call{RetrieveContext}{\mathcal{C},k}$
    \State $(a,\hat{c}) \gets
        f_{\mathrm{verify}}(d,u_k,t_k,S_m)$

    \If{$a=\textsc{Accept}$ \textbf{and}
        $\hat{c}\geq\tau_c$}
        \State $\hat{d} \gets
            \Call{RefineVerifiedEvent}{d,u_k,t_k,S_m}$
        \State $\mathcal{A} \gets
            \mathcal{A}\cup\{\hat{d}\}$
    \ElsIf{$a=\textsc{Merge}$}
        \State $\mathcal{Q} \gets
            \mathcal{Q}\cup\{(d,k,g)\}$
            \Comment{Resolve jointly}
    \Else
        \State \textbf{continue}
            \Comment{Discard rejected or low-confidence hypotheses}
    \EndIf
\EndFor

\State $\widehat{\mathcal{D}} \gets
    \Call{ConsolidateVerifiedEvents}
    {\mathcal{A},\mathcal{Q},\mathcal{D},\mathcal{C},
     \tau_{\mathrm{IoU}}}$
\State $R \gets f_{\mathrm{report}}(\widehat{\mathcal{D}})$
\State \Return $R$

\end{algorithmic}
\end{algorithm}
\section{Failure Schema}
\label{app:taxonomy}

\subsection{Scope, Construction and Status}
\label{app:schema_construction}

\paragraph{Scope.} The schema covers \emph{visible} failures in generated RGB
robot-manipulation video: any observable deviation of the video from a correct,
physically plausible execution of the instructed task, including rendering
conditions that make the execution visually undecidable. It deliberately does
not cover action executability on hardware, safety, or downstream policy harm,
and an empty report is a statement about the sampled frames rather than about
the depicted behaviour. Severity is likewise defined on observable video
properties, not on unmeasured downstream outcomes.

\paragraph{Construction.} We assembled the schema before production annotation
in three passes. First, top-down: we collected the evaluation criteria used by
twelve prior benchmarks and taxonomies: general video-quality
suites~\citep{huang2024vbench, zheng2025vbench, he2024videoscore},
physical-plausibility and physical-reasoning
benchmarks~\citep{paibench, zhang2025morpheus, zhang2026physion}, event-level
artifact detectors~\citep{taesiri2024glitchbench, zheng2026open,
gupta2026spotlight}, and robot-video
evaluators~\citep{deng2026rethinking, jiang2026robowm, qin2024worldsimbench},
and kept every criterion expressible as an observable property of a single
video. Second, bottom-up: two authors open-coded 120 generated clips
drawn from the three development generators but held out of the benchmark, writing free-text
descriptions of every failure and grouping them by the mechanism they named.
Third, reconciliation: over three rounds we merged criteria that no
annotator could separate in practice, split those that named two mechanisms,
and wrote an exclusion rule for every pair that had been confused during
the $30$-clip calibration pass (Appendix~\ref{app:annotation}). The result,
comprising definitions, exclusion rules and severity rubric, was frozen after
calibration and before any production clip was annotated, and was not revised
afterwards.

\paragraph{Design criteria.} Three requirements follow from the
setting rather than from taste, and they explain most of the schema's shape.
Every category must be decidable from sampled RGB frames alone, because that is
all an offline evaluator sees; where the frames do not decide, the schema has to
be able to say so instead of forcing a label, which is why three types are
evidence-limiting rather than assertions of defect. Neighbouring categories must
be separable in practice and not merely in principle, which is why every type
carries an exclusion rule naming the type that applies instead
(Appendix~\ref{app:codebook}), written for the pairs annotators confused during
calibration. And the vocabulary must be closed enough to aggregate over, yet
open enough to measure its own gaps: free-text description alone, as used by
open-ended detectors~\citep{zheng2026open}, gives no category to report
per-category reliability over and no way to state what the vocabulary fails to
cover.

\paragraph{On the uniform width.} Each dimension holding exactly five
types is a property of the reconciliation pass, not a finding about robot-video
failure. Merging criteria that no annotator could separate in calibration, and
splitting those that named two mechanisms, drove the dimensions toward a similar
granularity, and we fixed the width at that point so that the codebook could be
read and applied uniformly. The types are not equally natural and we do not
present them as such: per-type frequencies span $22$ to $138$ annotated events
(Table~\ref{tab:type_counts}), and one dimension, Visual Quality, carries most of the
evidence-limiting mass ($82$ of $113$ events). A schema chosen for balance rather than for the
phenomenon would show the opposite.

\paragraph{Status.} The schema is the annotation instrument of
\textsc{RoboGazeBench}, not a claim to an exhaustive account of robot-video
failure. We therefore report what an instrument should report: how often each
type is used (Table~\ref{tab:type_counts}), how reliably annotators apply it at
each level (Appendix~\ref{app:agreement_levels}), and what fraction of annotated
events it failed to cover (Appendix~\ref{app:residual}). Following standard
practice for taxonomy development~\citep{nickerson2013taxonomy} and for
empirically grounded failure taxonomies~\citep{cemri2025mast}, we treat coverage
as a measured quantity rather than an assertion.

\paragraph{Abstention and the residual type.} Three types are
\emph{evidence-limiting}: \texttt{ambiguous\_\allowbreak task\_\allowbreak success}, \texttt{occlusion}
and \texttt{low\_\allowbreak visibility} record that the sampled frames do not decide,
rather than asserting that the generator produced a defect. They account for
113 events (6.1\% of all annotated events). Keeping them
inside the schema makes abstention reportable instead of silent, and it is part
of why \emph{Visual Quality} has the lowest inter-annotator agreement of the six
dimensions (Appendix~\ref{app:agreement_levels}). One type is a \emph{residual}:
\texttt{other\_\allowbreak implausible\_\allowbreak dynamics} applies only when none of the four named
Physical Plausibility types does, and absorbs 13.7\% of Physical
Plausibility events. Neither category is scored differently; both are marked so
that a reader can separate what the schema asserts from what it declines to
decide.
\subsection{Type Definitions and Exclusion Rules}
\label{app:codebook}

Tables~\ref{tab:codebook_a} and~\ref{tab:codebook_b} give the released
codebook. Each type carries a positive definition and an explicit rule naming
the neighbouring type that applies instead; the pairs listed were the ones most
often confused during calibration. Annotators worked from this text verbatim.

\begin{table}[!h]
\centering
\scriptsize
\setlength{\tabcolsep}{3.4pt}
\renewcommand{\arraystretch}{1.08}
\caption{Codebook, Groups A and B (part 1): Task Semantics and the object
half of Entity Consistency.}
\label{tab:codebook_a}
\begin{tabular}{>{\raggedright\arraybackslash}p{0.175\textwidth}>{\raggedright\arraybackslash}p{0.365\textwidth}>{\raggedright\arraybackslash}p{0.375\textwidth}}
\toprule
\textbf{Type} & \textbf{Applies when} & \textbf{Does not apply when} \\
\midrule
\multicolumn{3}{@{}l}{\cellcolor{rgsummary}\textbf{1. Task Progress (TP)}: \emph{Does the task reach the instructed end state?}} \\
1.1 \texttt{task\_\allowbreak incompletion} & The task is attempted but the required final state is never reached. & The episode stops early with the action sequence unfinished ($\rightarrow$1.4), or a specific place action fails ($\rightarrow$1.3). \\
1.2 \texttt{failed\_\allowbreak grasp} & The robot attempts to grasp an object but never establishes a stable grasp. & The object follows the gripper with no contact ($\rightarrow$5.4), or a grasp succeeds and the placement fails ($\rightarrow$1.3). \\
1.3 \texttt{failed\_\allowbreak placement} & An object is transported or manipulated but not brought to its intended final state. & No stable grasp was formed ($\rightarrow$1.2), or the target itself is the wrong one ($\rightarrow$2.3). \\
1.4 \texttt{premature\_\allowbreak termination} & The episode ends before the task-relevant action sequence is complete. & The episode runs its full length without reaching the final state ($\rightarrow$1.1). \\
1.5 \texttt{ambiguous\_\allowbreak task\_\allowbreak success} & \emph{Evidence-limiting.} The final state is visually ambiguous, so success cannot be determined from the frames. & The frames do decide; then record the asserted type. If a nameable visual condition causes the ambiguity, also record 6.4 or 6.5. \\
\midrule
\multicolumn{3}{@{}l}{\cellcolor{rgsummary}\textbf{2. Instruction Consistency (IC)}: \emph{Does the execution follow the instruction?}} \\
2.1 \texttt{wrong\_\allowbreak effector} & The robot uses an effector inconsistent with the instruction or expected embodiment. & Extra, absent or duplicated limbs are rendered ($\rightarrow$4.1--4.3). \\
2.2 \texttt{wrong\_\allowbreak object} & The robot interacts with an object other than the one specified. & The correct object changes identity mid-clip ($\rightarrow$3.3). \\
2.3 \texttt{wrong\_\allowbreak target\_\allowbreak location} & The object is moved to a location inconsistent with the instruction. & The object reaches the correct target but is not stably placed ($\rightarrow$1.3). \\
2.4 \texttt{wrong\_\allowbreak action\_\allowbreak order} & Required actions occur in an order inconsistent with the instruction. & A completed action repeats with no ordering constraint stated ($\rightarrow$1.1). \\
2.5 \texttt{ignored\_\allowbreak constraint} & A stated constraint, precondition or task-specific requirement is ignored. & The violated constraint names an effector, object, location or order; use the specific type ($\rightarrow$2.1--2.4). \\
\midrule
\multicolumn{3}{@{}l}{\cellcolor{rgsummary}\textbf{3. Object--Scene Consistency (OSC)}: \emph{Do objects and the scene persist coherently?}} \\
3.1 \texttt{object\_\allowbreak hallucination} & A task-relevant object appears without a physically supported cause. & The object enters from outside the frame (normal), or a robot part appears ($\rightarrow$4.1). \\
3.2 \texttt{object\_\allowbreak disappearance} & A visible object vanishes without being moved out of view; requires visible evidence that occlusion cannot explain. & Occlusion is the better explanation ($\rightarrow$6.4). \\
3.3 \texttt{object\_\allowbreak identity\_\allowbreak swap} & An object changes semantic identity or is replaced by another object. & Only colour or texture changes ($\rightarrow$3.5); only geometry changes ($\rightarrow$3.4). \\
3.4 \texttt{object\_\allowbreak distortion} & Geometry deforms inconsistently with the object's expected rigidity or articulation. & Surface colour or texture changes with stable geometry ($\rightarrow$3.5); the deforming entity is the robot ($\rightarrow$4.4). \\
3.5 \texttt{object\_\allowbreak appearance\_\allowbreak drift} & Colour or texture drifts over time without task-relevant cause, geometry remaining stable. & Geometry changes ($\rightarrow$3.4); the whole frame degrades ($\rightarrow$6.1, 6.3). \\
\bottomrule
\end{tabular}
\end{table}

\begin{table}[!h]
\centering
\scriptsize
\setlength{\tabcolsep}{3.4pt}
\renewcommand{\arraystretch}{1.08}
\caption{Codebook (part 2): the robot half of Entity Consistency, and
Physics \& Observability.}
\label{tab:codebook_b}
\begin{tabular}{>{\raggedright\arraybackslash}p{0.175\textwidth}>{\raggedright\arraybackslash}p{0.365\textwidth}>{\raggedright\arraybackslash}p{0.375\textwidth}}
\toprule
\textbf{Type} & \textbf{Applies when} & \textbf{Does not apply when} \\
\midrule
\multicolumn{3}{@{}l}{\cellcolor{rgsummary}\textbf{4. Robot-Body Consistency (RBC)}: \emph{Does the robot keep its morphology and kinematics?}} \\
4.1 \texttt{hallucinated\_\allowbreak robot\_\allowbreak part} & A limb, gripper, finger or body segment appears with no physical continuity and no counterpart on the platform. & The part is an additional copy of one the platform does have ($\rightarrow$4.3). \\
4.2 \texttt{missing\_\allowbreak robot\_\allowbreak part} & A visible robot part becomes absent when it should remain visible. & The part leaves the frame or is occluded (not a failure; record 6.4 if it blocks diagnosis). \\
4.3 \texttt{duplicated\_\allowbreak arm\_\allowbreak or\_\allowbreak gripper} & An extra copy of a part the platform does have is rendered. & The part has no counterpart on the platform ($\rightarrow$4.1). \\
4.4 \texttt{robot\_\allowbreak body\_\allowbreak deformation} & The robot bends, stretches, melts or changes shape beyond its kinematic structure. & The deforming entity is an object ($\rightarrow$3.4). \\
4.5 \texttt{left\_\allowbreak right\_\allowbreak identity\_\allowbreak confusion} & Left/right arms, grippers or body sides swap identity or become inconsistent over time. & A side disappears rather than swapping ($\rightarrow$4.2). \\
\midrule
\multicolumn{3}{@{}l}{\cellcolor{rgsummary}\textbf{5. Physical Plausibility (PP)}: \emph{Is the depicted interaction physically supportable?}} \\
5.1 \texttt{object\_\allowbreak teleportation} & An object changes position discontinuously with no visible transport or causal interaction. & The jump occurs across an occluder and is consistent with some occluded path ($\rightarrow$6.4). \\
5.2 \texttt{object\_\allowbreak floating} & An object remains suspended with no visible support, contact or plausible dynamics. & The object follows the gripper without contact ($\rightarrow$5.4). \\
5.3 \texttt{object\_\allowbreak penetration} & A solid object passes through another solid object, robot part or support surface; requires an apparent intersection. & Contact is merely occluded and ambiguous ($\rightarrow$6.4). \\
5.4 \texttt{grasp\_\allowbreak without\_\allowbreak visible\_\allowbreak support} & An object follows or moves with the robot despite no visible grasp, contact or support. & The robot attempted a grasp and the object stayed put ($\rightarrow$1.2). \\
5.5 \texttt{other\_\allowbreak implausible\_\allowbreak dynamics} & \emph{Residual.} Motion violates continuity, gravity, contact or inertia in a way none of 5.1--5.4 names. & Any of 5.1--5.4 applies. Use only after all four have been excluded. \\
\midrule
\multicolumn{3}{@{}l}{\cellcolor{rgsummary}\textbf{6. Visual Quality (VQ)}: \emph{Is the video artifact-free and readable?}} \\
6.1 \texttt{frame\_\allowbreak corruption} & Frames contain rendering corruption, tearing, severe artifacts or missing visual content. & A localized geometry change affects one entity ($\rightarrow$3.4, 4.4). \\
6.2 \texttt{camera\_\allowbreak instability} & Camera shake, viewpoint jumps or unstable framing disrupt temporal interpretation. & The camera is stable and the scene layout itself changes ($\rightarrow$3.x). \\
6.3 \texttt{blur} & Motion or defocus blur prevents reliable interpretation of task-relevant state. & Frames are globally dark or low-contrast ($\rightarrow$6.5). \\
6.4 \texttt{occlusion} & \emph{Evidence-limiting.} Occlusion or masking prevents reliable observation of the relevant robot--object interaction. & The frames do decide; record the asserted failure type instead. \\
6.5 \texttt{low\_\allowbreak visibility} & \emph{Evidence-limiting.} Lighting, resolution or contrast is too poor for reliable diagnosis. & Blur is localized and lighting is adequate ($\rightarrow$6.3). \\
\bottomrule
\end{tabular}
\end{table}

\FloatBarrier

\subsection{Severity, Cascading Events and Primary Dimension}
\label{app:severity}

\paragraph{Severity rubric.}
Severity is an ordinal score in $\{1,\dots,5\}$ defined on observable task-video
properties rather than unmeasured downstream outcomes:
$1$ cosmetic (task state and object interaction remain interpretable);
$2$ minor (localized, little ambiguity about the intended execution);
$3$ moderate (affects an object, robot part or subtask transition relevant to the
task);
$4$ severe (obscures whether the task executed correctly, or makes the clip
unsuitable as a clean demonstration);
$5$ catastrophic (a physically or semantically impossible event that invalidates
the execution). Severe-event recall counts events at severity $4$ or $5$;
these are 558 events, 30.2\% of the total
(Table~\ref{tab:type_counts}).

\paragraph{Cascading events.}
An event may implicate several dimensions: a gripper passing through a cup is
both physical-plausibility and robot-body. Annotators assign all applicable
dimensions and promote a primary one from the most directly observable evidence;
matching uses the primary dimension and the manifest retains the rest.
Cascading events are kept rather than discarded: 163 events
across $114$ clips ($12.5\%$) implicate more than one
dimension. All per-type counts below are over primary labels, so they sum to the
event total. The label-agnostic setting $\lambda_{\rm dim}=0$
(Appendix~\ref{app:robustness}) discards the dimension entirely and is reported
as a robustness check.

\subsection{Per-Type Event Counts}
\label{app:type_counts}

Across the $914$ clips, annotators recorded 1{,}847 failure events on
704 clips (77.0\%); the remaining 210 clips
(23.0\%) carry no annotated event and form the failure-free set used for
clean accuracy. A clip with at least one event carries 2.6 on
average and 2 at the median.
Table~\ref{tab:type_counts} gives the distribution over the thirty types. No
type is unused, and the most frequent type (\texttt{task\_\allowbreak incompletion}, $138$) is
roughly six times the least frequent (\texttt{ignored\_\allowbreak constraint}, $22$); we report
the raw counts rather than a balanced sample because the distribution is a
property of the generators, not a design choice.
\begin{table}[!h]
\centering\scriptsize
\setlength{\tabcolsep}{4pt}
\renewcommand{\arraystretch}{1.05}
\caption{Annotated failure events per type, over primary labels, all
$914$ clips. Dimension totals sum to 1{,}847.}
\label{tab:type_counts}
\begin{tabular}{@{}llr@{\hspace{1.2em}}llr@{}}
\toprule
\textbf{Type} & & $n$ & \textbf{Type} & & $n$ \\
\midrule
\multicolumn{3}{@{}l}{\cellcolor{rgsummary}\textbf{1. Task Progress} \hfill 412} & \multicolumn{3}{l}{\cellcolor{rgsummary}\textbf{4. Robot-Body Cons.} \hfill 286} \\
1.1 & task incompletion              & 138   & 4.1 & hallucinated robot part      & 74 \\
1.2 & failed grasp                   & 104   & 4.2 & missing robot part           & 58 \\
1.3 & failed placement               & 87 & 4.3 & duplicated arm or gripper    & 47 \\
1.4 & premature termination          & 52  & 4.4 & robot body deformation       & 81 \\
1.5 & ambiguous task success & 31 & 4.5 & left/right identity confusion & 26 \\
\midrule
\multicolumn{3}{@{}l}{\cellcolor{rgsummary}\textbf{2. Instruction Cons.} \hfill 198} & \multicolumn{3}{l}{\cellcolor{rgsummary}\textbf{5. Physical Plausibility} \hfill 379} \\
2.1 & wrong effector        & 28   & 5.1 & object teleportation                 & 86 \\
2.2 & wrong object          & 61   & 5.2 & object floating                      & 71 \\
2.3 & wrong target location & 54 & 5.3 & object penetration                   & 112 \\
2.4 & wrong action order    & 33  & 5.4 & grasp without visible support        & 58 \\
2.5 & ignored constraint    & 22  & 5.5 & other implausible dynamics & 52 \\
\midrule
\multicolumn{3}{@{}l}{\cellcolor{rgsummary}\textbf{3. Object--Scene Cons.} \hfill 341} & \multicolumn{3}{l}{\cellcolor{rgsummary}\textbf{6. Visual Quality} \hfill 231} \\
3.1 & object hallucination    & 63   & 6.1 & frame corruption          & 44 \\
3.2 & object disappearance    & 98   & 6.2 & camera instability        & 38 \\
3.3 & object identity swap    & 41 & 6.3 & blur                      & 67 \\
3.4 & object distortion       & 79  & 6.4 & occlusion                 & 49 \\
3.5 & object appearance drift & 60  & 6.5 & low visibility            & 33 \\
\midrule
\rowcolor{rgsummary}
\multicolumn{3}{@{}l}{\textbf{Total events}} & \multicolumn{3}{r}{\textbf{1{,}847}} \\
\multicolumn{3}{@{}l}{Severity 1 / 2 / 3} & \multicolumn{3}{r}{329 / 448 / 512} \\
\multicolumn{3}{@{}l}{Severity 4 / 5 \quad (30.2\% of events)} & \multicolumn{3}{r}{388 / 170} \\
\bottomrule
\end{tabular}
\end{table}

\FloatBarrier

\subsection{Events the Schema Did Not Cover}
\label{app:residual}

Annotators could mark an event \texttt{unresolved} when no type applied, so that
a failure outside the schema would be recorded rather than forced into the
nearest label. This occurred on 73 events (4.0\% of all
annotated events) across $61$ clips ($6.7\%$).

A senior adjudication pass re-examined every unresolved event against the
codebook and sorted it into one of three outcomes. 52 of them
(71\%) resolved into an existing type once the existing boundary rule
between the two candidates was applied explicitly, almost all on the
\texttt{object\_\allowbreak distortion} / \texttt{object\_\allowbreak appearance\_\allowbreak drift} and
\texttt{failed\_\allowbreak grasp} / \texttt{grasp\_\allowbreak without\_\allowbreak visible\_\allowbreak support} pairs. These
were application errors rather than coverage gaps: the exclusion rules in
Tables~\ref{tab:codebook_a} and~\ref{tab:codebook_b} already decide them, and
the frozen codebook was not changed.
9 were re-labelled as normal execution or annotator error and
removed. 12 (16\% of unresolved, 0.6\% relative to the 1{,}847 typed
events) remained outside the schema; they are released separately under an
\texttt{other} category together with their free-text descriptions and are not
counted in Table~\ref{tab:type_counts} or in any score. Their
recurring kinds are illumination and shadow inconsistency not attached to any
entity (5), behaviour of background agents who are not the
manipulating robot (4), and 3 singletons.

We report this residual rather than suppressing it. A schema with no escape
hatch cannot measure its own coverage, so a rate of exactly zero would be
evidence about the annotation interface rather than about the schema; a measured
0.6\% is what a reader needs in order to know how much of the phenomenon
the thirty types account for. Illumination and shadow inconsistency is the
clearest candidate for a type in a future revision, and it recurs in the
held-out coverage check below.
\subsection{Coverage Check on Held-Out Generators}
\label{app:coverage}

The residual above is measured by annotators trained on the schema, so it is
anchored by construction: a failure the schema does not name is also one they
were not taught to look for. We therefore ran a second, schema-blind check on
the held-out block. On 40 clips from Wan~2.2 and Veo, neither of which was
used when the schema was frozen or the evaluator configured
(Appendix~\ref{app:contamination}), annotators described every failure in
free text without seeing the schema; a separate annotator then mapped each
description to the thirty types.

Of 96 described failures, 81 (84\%) mapped to
exactly one type, 11 (11\%) required two or more (cascading
events, which the primary-dimension rule already handles), and 4\
(4\%) could not be mapped, giving 96\% coverage on
generators the schema was not built against. One candidate type
recurred across the unmapped descriptions, the same illumination and shadow
inconsistency found in the residual analysis. We read this as the schema
transferring to unfamiliar generators with a small, named and reproducible gap,
rather than as completeness.
\FloatBarrier

\section{Benchmark Construction and Annotation}
\label{app:benchmark}

\subsection{Construction Overview}
\label{app:construction_overview}

\begin{figure}[!h]
\centering
\includegraphics[width=\linewidth]{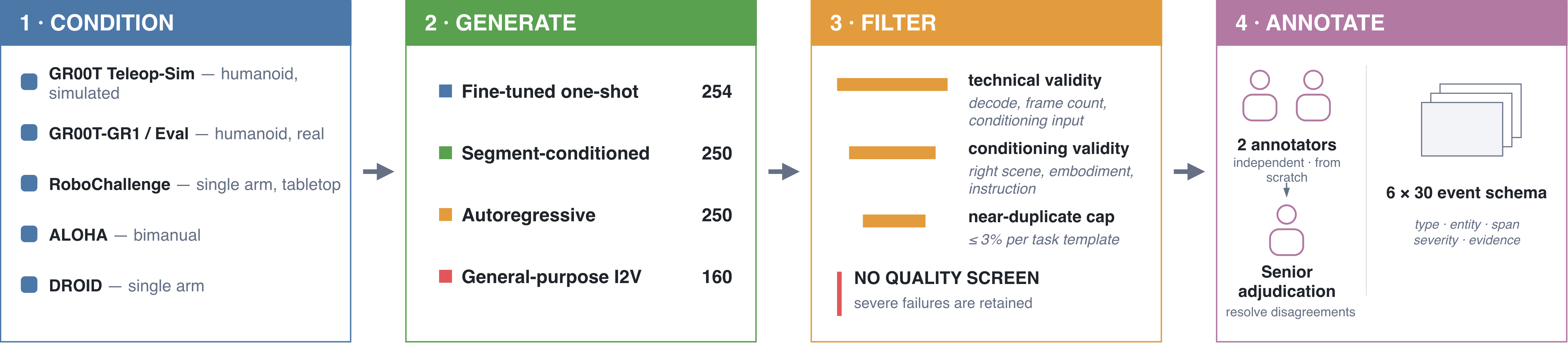}
\caption{Construction of \textsc{RoboGazeBench}. Conditioning instructions
and initial observations are drawn from five source datasets, rendered by five
generators under four generation mechanisms, filtered for technical and
conditioning validity but never for output quality, and annotated from scratch
under the frozen event schema.}
\label{fig:construction}
\end{figure}

\subsection{Composition}
\label{app:composition}

\begin{figure}[!h]
\centering
\includegraphics[width=0.92\linewidth]{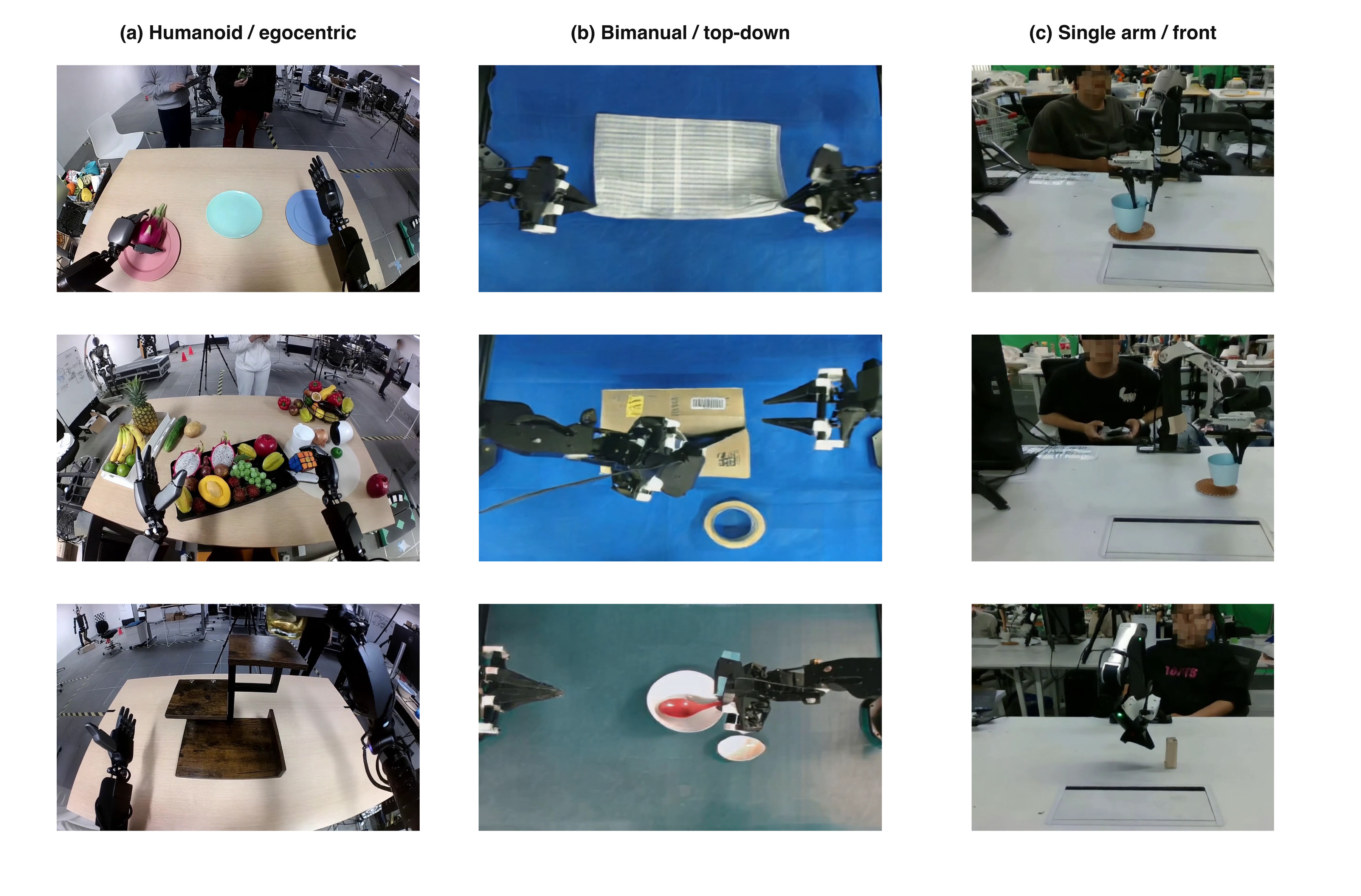}
\caption{Example frames from \textsc{RoboGazeBench}, three per group:
(a)~humanoid with egocentric view, (b)~bimanual with top-down view, and
(c)~single-arm with front view. Camera view is fixed by each source's capture rig
and therefore coincides with embodiment.}
\label{fig:dataset_examples}
\end{figure}

Table~\ref{tab:composition} gives the counts behind
Figure~\ref{fig:composition}, together with camera view and resolution, which the
figure does not show. Example frames for the three embodiment and camera-view groups are
shown in Figure~\ref{fig:dataset_examples}. Humanoid clips are the largest embodiment group because the
fine-tuned one-shot block is entirely GR1; simulation-derived conditioning comes
from a single source inside that block, which is why the environment axis is
unbalanced.

\begin{table}[!h]
\centering\small
\caption{Counts along every reported axis. Each axis sums to $914$.
\emph{Embodiment} is defined by the arms a clip uses, not by the platform, so a
two-armed platform performing a one-armed task counts as single-arm.
\emph{Environment} describes the data a clip's conditioning came from; every clip
is generated. \emph{Camera view} is fixed by each source's capture rig and
therefore coincides with embodiment.}
\label{tab:composition}
\begin{minipage}[t]{0.49\linewidth}
\centering
\setlength{\tabcolsep}{5pt}
\begin{tabular}{@{}llr@{}}
\toprule
Axis & Group & $n$ \\
\midrule
\multirow{4}{*}{Mechanism}
 & Fine-tuned one-shot     & 254 \\
 & Segment-conditioned     & 250 \\
 & Autoregressive          & 250 \\
 & General-purpose I2V     & 160 \\
\midrule
\multirow{5}{*}{Generator}
 & Cosmos-Predict2.5       & 254 \\
 & DreamGen                & 250 \\
 & DreamDojo               & 250 \\
 & Wan 2.2                 &  80 \\
 & Veo                     &  80 \\
\midrule
\multirow{3}{*}{Embodiment}
 & Humanoid                & 414 \\
 & Bimanual                & 250 \\
 & Single arm              & 250 \\
\midrule
\multirow{2}{*}{Environment}
 & Real-derived            & 760 \\
 & Simulation-derived      & 154 \\
\bottomrule
\end{tabular}
\end{minipage}
\hfill
\begin{minipage}[t]{0.49\linewidth}
\centering
\setlength{\tabcolsep}{5pt}
\begin{tabular}{@{}llr@{}}
\toprule
Axis & Group & $n$ \\
\midrule
\multirow{4}{*}{Horizon}
 & $\leq 15$\,s            & 619 \\
 & $15$--$40$\,s           & 165 \\
 & $40$--$80$\,s           &  95 \\
 & $>80$\,s                &  35 \\
\midrule
\multirow{4}{*}{Task family}
 & T1 free-space transport & 567 \\
 & T2 constrained contact  &  81 \\
 & T3 articulated \& activ.& 155 \\
 & T4 coord.\ \& deformable& 111 \\
\midrule
\multirow{3}{*}{Camera view}
 & Egocentric              & 414 \\
 & Top-down                & 250 \\
 & Front                   & 250 \\
\midrule
\multirow{5}{*}{Resolution}
 & $1280{\times}800$       & 254 \\
 & $1280{\times}720$       &  80 \\
 & $832{\times}480$        &  80 \\
 & $768{\times}432$        & 155 \\
 & $640{\times}480$        & 345 \\
\bottomrule
\end{tabular}
\end{minipage}
\end{table}

\FloatBarrier

\subsection{Generation Settings}
\label{app:gensettings}

Three of the four mechanisms are realised on Cosmos-family backbones:
Cosmos-Predict2.5-14B fine-tuned per task, Cosmos-Predict2 2B driven by DreamGen,
and a 2B action-conditioned DiT for DreamDojo. \textsc{RoboGazeBench} therefore
varies \emph{generation mechanism} more than it varies backbone architecture, and
Wan~2.2 and Veo are the only architecturally independent systems in it. We state
this rather than let the pipeline names imply otherwise; the mechanism contrast
is partly controlled because DreamGen and DreamDojo share source datasets and
a backbone family, and architectural independence is what the held-out block
supplies. Clip duration is set by each generator's rollout mechanism rather than
by the condition alone: segment-conditioned DreamGen renders bounded segments,
whereas autoregressive DreamDojo continues for as long as its conditioning
actions run, which is why the two blocks differ in horizon
(Table~\ref{tab:generators}). Resolution and frame rate also differ by source, so a fixed frame
budget corresponds to a different real-time sampling interval in each block;
Appendix~\ref{app:robustness} reports the result at half, the default, and
double each backbone's frame budget.

\begin{table}[!h]
\centering\scriptsize
\setlength{\tabcolsep}{4.5pt}
\renewcommand{\arraystretch}{1.18}
\caption{Per-source generation settings, grouped by generation mechanism.
Each shaded header names the backbone and the sampler setting (classifier-free
guidance / denoising steps) shared by every source under it. Durations are
mean\,[min--max] where the horizon varies; frame counts are duration $\times$ FPS
and are given per clip in the release. $^{\dagger}$Aggregate over the two sources
above it.}
\label{tab:generators}
\begin{tabular}{@{}>{\raggedright\arraybackslash}p{0.235\linewidth}llc>{\raggedright\arraybackslash}p{0.175\linewidth}r@{}}
\toprule
Source & Embodiment & Resolution & FPS & Duration\,(s) & $n$ \\
\midrule
\rowcolor{rgsummary}
\multicolumn{6}{@{}l}{\textbf{Fine-tuned one-shot}\quad Cosmos-Predict2.5-14B\quad 7 / 35} \\
\emph{ft.\ GR00T Teleop-Sim}        & humanoid   & $1280{\times}800$ & 16 & 8.69  & 154 \\
\emph{ft.\ GR00T-GR1 + Eval}        & humanoid   & $1280{\times}800$ & 16 & 5.80  & 100 \\
\midrule
\rowcolor{rgsummary}
\multicolumn{6}{@{}l}{\textbf{Segment-conditioned} (DreamGen)\quad Cosmos-Predict2 2B\quad 7 / 35, seed 0} \\
\texttt{sa\_static\_v2/\hspace{0pt}90k}    & single arm & $640{\times}480$ & 16 & 23.06\,[11.6--40.3]  & 95 \\
\texttt{bimanual\_16fps/\hspace{0pt}110k}  & bimanual   & $768{\times}432$ & 16 & 63.31\,[11.6--103.6] & \\
\texttt{humanoid\_16fps/\hspace{0pt}90k}   & humanoid   & $768{\times}432$ & 16 & 14.31\,[11.6--17.3]  & $155^{\dagger}$ \\
\midrule
\rowcolor{rgsummary}
\multicolumn{6}{@{}l}{\textbf{Autoregressive} (DreamDojo)\quad action-conditioned DiT 2B\quad 1.5 / 35} \\
\texttt{single\_arm\_sv/\hspace{0pt}30k}   & single arm & $640{\times}480$ & 15 & 12.47\,[8.9--38.5]   & 100 \\
\texttt{aloha\_resume70k\_\hspace{0pt}bimanual\_sv/\hspace{0pt}100k} & bimanual & $640{\times}480$ & 10 & 49.30\,[12.1--148.9] & \\
\texttt{2B\_GR1\_post-\hspace{0pt}train/50k} & humanoid & $640{\times}480$ & 10 & 9.80\,[4.9--142.0]   & $150^{\dagger}$ \\
\midrule
\rowcolor{rgsummary}
\multicolumn{6}{@{}l}{\textbf{General-purpose I2V}\quad independent backbones, no robot-specific training} \\
Wan2.2-I2V-14B                      & all three  & $832{\times}480$  & 16 & 5.06 & 80 \\
Veo                                 & all three  & $1280{\times}720$ & 24 & 8.00 & 80 \\
\midrule
\rowcolor{rgsummary}
\multicolumn{5}{@{}l}{\textbf{Total}} & \textbf{914} \\
\bottomrule
\end{tabular}
\end{table}

\begin{figure}[!h]
\centering
\includegraphics[width=0.92\linewidth]{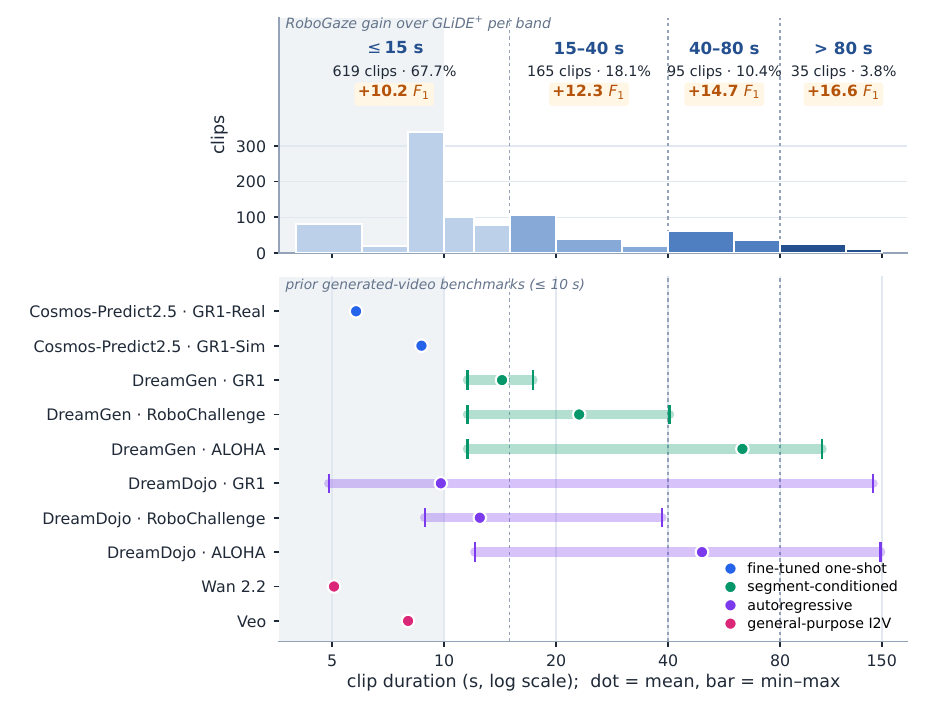}
\caption{Horizon of \textsc{RoboGazeBench}. \emph{Top:} clip-duration
distribution coloured by horizon band, with each band's exact clip count and
\textsc{RoboGaze}'s gain over \gplus\ in that band printed above it.
\emph{Bottom:} the range and mean duration of each source, coloured by generation
mechanism, on the same axis. The
shaded region is the horizon at which prior generated-video benchmarks evaluate;
only the robot-specific world models produce clips beyond it by design, which is
why the held-out-generator and long-horizon analyses in
Section~\ref{sec:results} are reported separately. Within-bin shape reproduces
the reported quantiles.}
\label{fig:horizon}
\end{figure}

\FloatBarrier

\subsection{Sampling and Matched Design}
\label{app:sampling}

Conditions were sampled before any generated output was viewed, so the benchmark
cannot have been selected for failure density. Within each generator the sample is
stratified over the source's task families and over its frame-count range. The
second of these matters more than it appears: the bimanual sources are
intrinsically the longest (Figure~\ref{fig:horizon}), so without stratifying on
duration inside each source, the horizon analysis in Section~\ref{sec:horizon}
would be indistinguishable from an embodiment effect.

Wan~2.2 and Veo render the same $80$ conditions. The other generators draw their
conditions from overlapping source datasets but were sampled separately, by task
family and duration, so their conditions are not paired one-to-one. That $914$
clips draw on only $194$ instructions reflects that one instruction is realised
under several distinct initial observations and by several generators. Clips
sharing a condition share a resampling group in the bootstrap
(Appendix~\ref{app:metrics}), so repeated conditions do not inflate the effective
sample size.
\subsection{Filtering}
\label{app:filtering}

A clip is rejected on two grounds only. It is technically invalid (a decode
failure, a truncated file, a frame count outside the declared range, a missing
conditioning input), in which case it is regenerated once with the same seed and
dropped if it fails again. Or its conditioning is wrong: the clip does not
depict the instructed scene, embodiment or task, judged by an annotator who sees
the instruction and the first frame only. Near-duplicate conditions are detected
by instruction and initial-frame similarity, and no task template exceeds $3\%$
of the benchmark.

Nothing is removed for poor output. A generated video that teleports an object,
deforms the gripper or abandons the task is what the benchmark exists to contain,
and screening on quality would yield a benchmark on which every evaluator looks
better than it is. One exception is historical: the $254$ fine-tuned one-shot
clips were quality-screened when first collected, before this policy was fixed.
The screen makes them easier than the rest, so every result is reported per
generator group (Appendix~\ref{app:slices}) where its effect is visible.
\subsection{Released Record Format}
\label{app:record}

Clip fields are fixed at construction and event fields are produced by
annotation and adjudication; Table~\ref{tab:record} shows one record of each. A
clip carrying no event is a \emph{clean} clip and contributes only to clean
accuracy.

\begin{table}[!h]
\centering\scriptsize
\setlength{\tabcolsep}{4pt}
\caption{A single \textsc{RoboGazeBench} record, abbreviated. Each clip
carries zero or more events; a clip with none is a \emph{clean} clip and
contributes to clean accuracy.}
\label{tab:record}
\begin{tabular}{@{}>{\raggedright\arraybackslash}p{0.26\linewidth}>{\raggedright\arraybackslash}p{0.70\linewidth}@{}}
\toprule
\multicolumn{2}{@{}l}{\textit{Clip}} \\
\texttt{clip\_id}            & \texttt{rgb\_00417} \\
\texttt{instruction}         & \emph{``pick up the glass and place it on the right tray''} \\
\texttt{condition\_id}       & \texttt{aloha\_sv\_0132} \\
\texttt{generator}           & \texttt{dreamgen} \quad$\vert$\quad \texttt{mechanism: segment\_conditioned} \\
\texttt{checkpoint}          & \texttt{bimanual\_16fps/110k} \\
\texttt{source\_dataset}     & \texttt{aloha} \quad$\vert$\quad \texttt{environment: real\_derived} \\
\texttt{embodiment}          & \texttt{bimanual} \\
\texttt{task\_family}        & \texttt{T1\_free\_space\_transport} \quad$\vert$\quad \texttt{fine\_task: object\_transport} \\
\texttt{camera\_view}        & \texttt{top\_down} \\
\texttt{resolution / fps}    & \texttt{768x432} / \texttt{16} \\
\texttt{frames / duration\_s}& \texttt{712} / \texttt{44.50} \quad$\vert$\quad \texttt{duration\_band: long\_40\_80s} \\
\texttt{seed}                & \texttt{0} \\
\texttt{filter\_status}      & \texttt{passed} \quad$\vert$\quad \texttt{regenerated: false} \\
\midrule
\multicolumn{2}{@{}l}{\textit{Event} \textnormal{(1 of 2 on this clip)}} \\
\texttt{event\_id}           & \texttt{rgb\_00417\_e1} \\
\texttt{dimension}           & \texttt{physical\_plausibility} \quad$\vert$\quad \texttt{secondary: [object\_scene]} \\
\texttt{type}                & \texttt{object\_penetration} \\
\texttt{entity}              & \texttt{glass} \\
\texttt{span\_frames / span\_s} & \texttt{[498, 546]} / \texttt{[31.1, 34.1]} \\
\texttt{severity}            & \texttt{4} \\
\texttt{description}         & \emph{``the left gripper closes through the glass rim instead of around it; the glass follows the gripper with no contact''} \\
\texttt{evidence}            & \emph{``frames 502--538, left gripper and glass rim''} \\
\texttt{flags}               & \texttt{open\_world: false} \quad$\vert$\quad \texttt{ambiguous: false} \\
\texttt{annotation}          & \texttt{annotators: [a3, a5]} \quad$\vert$\quad \texttt{adjudicated\_by: s1} \quad$\vert$\quad \texttt{double\_labelled: true} \\
\bottomrule
\end{tabular}
\end{table}

\FloatBarrier

\subsection{Annotation Protocol and Interface}
\label{app:annotation}

Seven annotators with bachelor-level robotics, computer-vision or
mechanical-engineering backgrounds completed a four-hour rubric training session,
a $30$-clip calibration pass and a senior review before production; three were
designated senior on calibration agreement above $\kappa=0.78$. Each clip is
annotated independently from scratch by two annotators and merged by a senior
annotator under a written adjudication rubric: detection disagreements are
re-watched and resolved, dimension disagreements default to the cascade rule,
spans with pairwise $\text{IoU}\ge0.6$ are merged by union and otherwise treated
as separate events, and severity disagreements within $\pm1$ are averaged, otherwise re-rated from
scratch; the manifest keeps the averaged value, and event counts and
severe-event recall use it rounded up to the more severe level.
Annotators see only the generated video and the task instruction, never a dataset
identifier, a generator checkpoint or a model output, and the schema, rubric
and guide were frozen before production annotation. Annotators were paid above
local hourly research-assistant rates; the study uses generated video only and
was reviewed internally as non-human-subject work.

\begin{figure}[!h]
\centering
\includegraphics[width=0.92\linewidth]{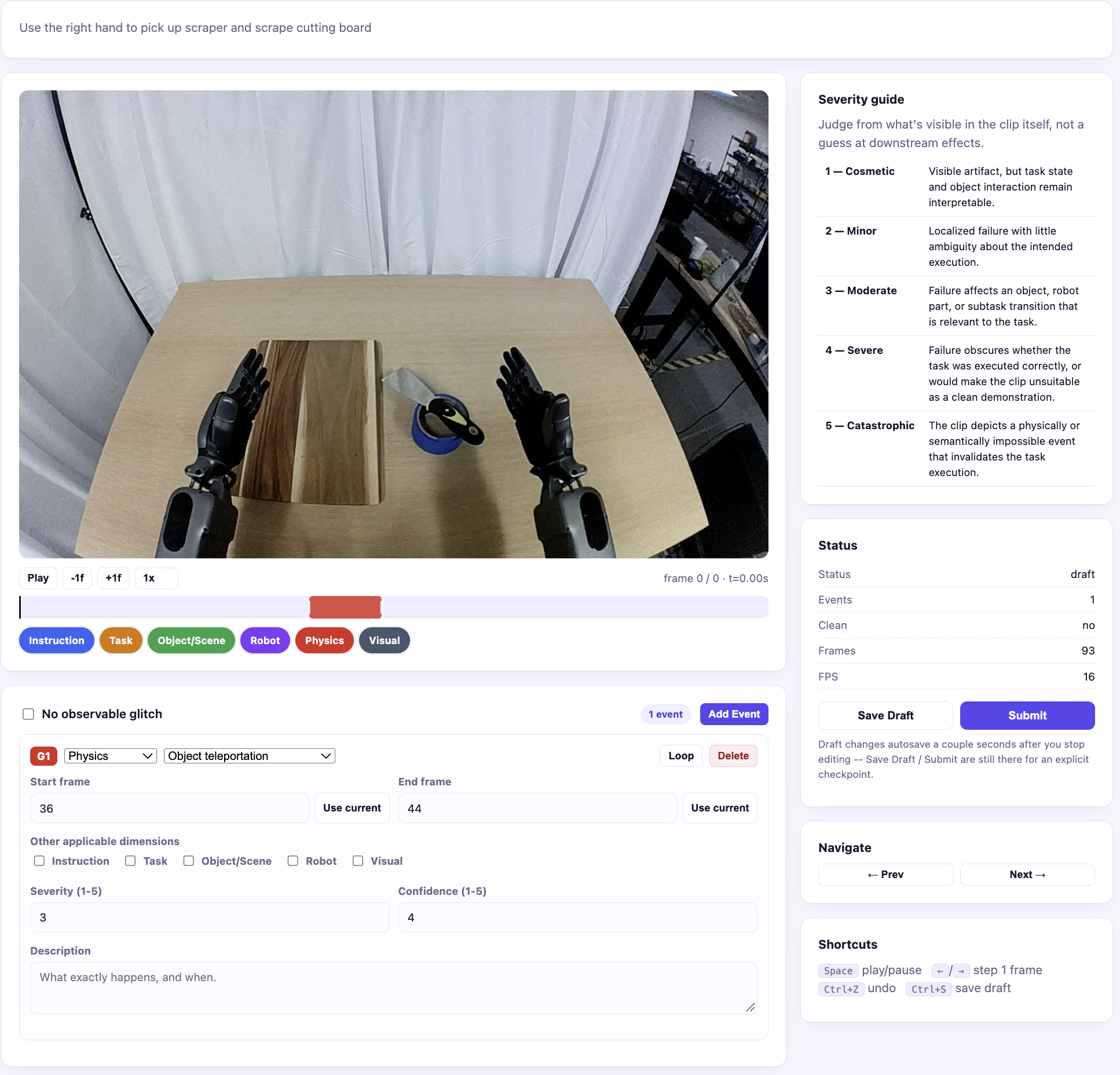}
\caption{The annotation interface. Annotators see the generated video and the
task instruction only.}
\label{fig:annotation_ui}
\end{figure}

\subsection{Annotation Effort}
\label{app:effort}

Event annotation is slow because a clip must be watched more than once before a
span can be fixed: an annotator views it for gist, again for candidate events,
then scrubs each candidate at reduced speed to set boundaries at $0.1$\,s
precision before writing the description and severity. A pass therefore costs
roughly twenty-four times real time, and it scales with clip length rather than
with clip count: a minute-long clip takes five to six times as long as a
ten-second one. Table~\ref{tab:effort} accounts for the whole labelling effort.

\textsc{RoboGazeBench} took approximately $456$ annotator-hours, of which $245$
were production passes and $49$ adjudication. The median clip cost $5$ minutes
per pass and the mean $8$, the gap between the two being entirely the long tail
of the horizon distribution (Figure~\ref{fig:horizon}). We report this because
the cost of the benchmark is dominated by the property that distinguishes it:
nearly half of the total labelling time went to the $14\%$ of clips that run
beyond forty seconds, because those clips carry close to half of the
benchmark's total duration.
\begin{table}[!h]
\centering\small
\setlength{\tabcolsep}{6pt}
\caption{Labelling effort. Passes are independent annotations of the same
clips.}
\label{tab:effort}
\begin{tabular}{@{}llrr@{}}
\toprule
Stage & Scope & Passes & Hours \\
\midrule
Training and calibration   & 7 annotators, 30-clip calibration & --   & 49 \\
Production annotation      & all 914 clips                     & 2    & 245 \\
Adjudication               & all 914 clips                     & 1    & 49 \\
Gold subset                & 180 clips, senior annotators      & 3    & 72 \\
Blinded human audit        & 180 clips, two methods            & 3    & 41 \\
\midrule
\rowcolor{rgsummary}
\textbf{Total} & & & \textbf{456} \\
\bottomrule
\end{tabular}
\end{table}

\subsection{Contamination}
\label{app:contamination}

A benchmark assembled from existing video has to argue that the evaluated models
have not already seen it. \textsc{RoboGazeBench} does not have that problem in
its usual form, because every clip was generated for this work and none has been
published: no evaluated backbone can have encountered a benchmark clip, and the
failure events are properties of those clips, annotated here for the first time.

Two weaker channels remain. The conditioning instructions and initial
observations are drawn from public datasets, so a backbone may recognise a scene
layout or a task phrasing; this could help it describe what a clip is attempting
but not whether the attempt went wrong, which is what the benchmark scores.
And the schema and annotation guide are released with the benchmark, so models
trained after release may have seen the vocabulary. We report exact API snapshot
dates and checkpoint identifiers (Table~\ref{tab:model_snapshots}) so that either
channel can be checked against any model a later reader evaluates.

One block settles the question by construction. The Wan~2.2 and Veo clips were
generated after every model in Table~\ref{tab:main_results} was released and after
the first access dates in Table~\ref{tab:model_snapshots}, so no model could have
been trained on them, their conditioning renderings, or their annotations. Those are also the two generators on which
\textsc{RoboGaze}'s margin is largest (Appendix~\ref{app:slices}). Whatever
residual contamination the public conditioning sources introduce, it does not
account for the result. For the closed APIs we have no visibility into training
data and do not claim any; this is a limitation of evaluating proprietary models
rather than of the benchmark.
\subsection{Agreement}
\label{app:agreement}

Because every clip receives two independent passes, agreement is measured
over the whole benchmark rather than a subset. Deciding
\emph{whether} a clip contains a failure is markedly more reliable than agreeing
on the span of a physical-plausibility event ($\kappa=.74$ against a temporal IoU
of $.53$), and agreement falls monotonically with horizon, reaching on the
longest clips ($>80$\,s) about the level of the hardest dimensions. Across
dimensions, the ordering closely follows the one the models produce
(Figure~\ref{fig:agreement_gap}).

\begin{table}[!h]
\centering\small
\setlength{\tabcolsep}{4.5pt}
\caption{Inter-annotator agreement over all $914$ clips, each independently
annotated twice. $\kappa$ and $\alpha$ count an event that the other annotator
did not mark as a disagreement; temporal IoU is between the spans of temporally
matched events of the two production passes. These are raw agreement statistics
and are not the \emph{human agreement} reference of Section~\ref{sec:results},
which scores each gold-subset senior annotator under the model matching protocol
(Section~\ref{sec:protocol}).}
\label{tab:agreement}
\begin{tabular}{lccccc}
\toprule
 & Detection $\kappa$ & Dimension $\kappa$ & Temporal IoU & Severity $\alpha$ & Desc.\ sim. \\
\midrule
\rowcolor{rgsummary}
\textbf{Overall}        & \textbf{.74} & \textbf{.66} & \textbf{.61} & \textbf{.58} & \textbf{.73} \\
\midrule
$\leq 15$\,s            & .77 & .69 & .64 & .61 & .75 \\
$15$--$40$\,s           & .71 & .62 & .56 & .55 & .70 \\
$40$--$80$\,s           & .67 & .60 & .54 & .51 & .68 \\
$>80$\,s                & .63 & .56 & .50 & .46 & .64 \\
\midrule
Task Progress           & .79 & --  & .64 & .61 & .76 \\
Instruction Consistency & .81 & --  & .67 & .63 & .79 \\
Object--Scene Cons.     & .72 & --  & .60 & .57 & .72 \\
Robot-Body Cons.        & .70 & --  & .59 & .56 & .70 \\
Physical Plausibility   & .63 & --  & .53 & .50 & .65 \\
Visual Quality          & .58 & --  & .49 & .47 & .62 \\
\bottomrule
\end{tabular}
\end{table}

\subsection{Agreement at Three Granularities}
\label{app:agreement_levels}

Table~\ref{tab:agreement} measures agreement on whether a clip contains a
failure and on the dimension of an event. Because the schema also has a type
level, we report agreement there as well, on the $180$-clip gold subset where
three senior annotators labelled every clip independently. Events are first
matched temporally between annotators. The $\kappa$ and $\alpha$ columns are then
computed over all annotated events, with an event that no other annotator matched
counted as a disagreement; the two hierarchical columns are computed over matched
pairs only. The same-dimension rate ($84\%$) is therefore higher than
$\kappa_{\text{dim}}$ alone would imply, because it conditions on the event having
been found by both annotators.

Reliability falls monotonically with granularity: $.76$ for detection,
$.68$ for the dimension, and $.58$ for the exact type
(Krippendorff's $\alpha=.60$ over the thirty categories). Read
hierarchically, 84\% of matched pairs agree on the dimension and
60\% on the exact type, so most type-level disagreement is a choice
between siblings rather than a different reading of the event. This ordering is
the reason the main paper aggregates at the dimension level
(Table~\ref{tab:main_results}) and uses the type level only inside the
specialists and the released manifest: the dimension level is where annotators
agree well enough for a benchmark score, and the type level is where a
diagnostic report is informative enough to act on.

Three sibling pairs account for about half of all type-level disagreement:
\texttt{object\_distortion} against \texttt{object\_appearance\_drift}
(31 disagreeing matched pairs), \texttt{hallucinated\_robot\_part}
against \texttt{duplicated\_arm\_or\_gripper} (24), and
\texttt{object\_teleportation} against \texttt{other\_implausible\_dynamics}
(19). All three are the pairs whose exclusion rules were written during
calibration (Appendix~\ref{app:codebook}); they are boundary choices within a
dimension rather than disagreements about what happened in the clip.

Per dimension, exact-type agreement follows the same ordering as detection
agreement (\emph{Instruction Consistency} $.66$, \emph{Task Progress}
$.61$, \emph{Object--Scene} $.59$, \emph{Robot-Body} $.56$,
\emph{Physical Plausibility} $.51$, \emph{Visual Quality} $.49$),
and the two weakest are Visual Quality, which holds two of the three
evidence-limiting types, and Physical Plausibility, which holds the residual
type. What sampled RGB frames can settle bounds the schema in the same
place it bounds the evaluator (Figure~\ref{fig:agreement_gap}).
\begin{table}[!h]
\centering\small
\setlength{\tabcolsep}{5pt}
\caption{Agreement at three granularities, $180$-clip gold subset,
three independent senior annotations per clip. The $\kappa$ and $\alpha$ columns
are computed over all annotated events, an unmatched event counting as a
disagreement; the hierarchical columns are computed over temporally matched event
pairs only.}
\label{tab:agreement_levels}
\begin{tabular}{lcccc}
\toprule
Granularity & $\kappa$ & $\alpha$ & \multicolumn{2}{c}{Hierarchical (matched pairs)} \\
\cmidrule(lr){4-5}
 & & & same dimension & same type \\
\midrule
Clip-level detection & .76 & -- & -- & -- \\
Event dimension (6)  & .68 & -- & 84\% & -- \\
\rowcolor{rgaccent}
Event type (30)      & .58 & .60 & 84\% & 60\% \\
\midrule
\multicolumn{5}{@{}l}{\cellcolor{rgsummary}\textit{Exact-type $\kappa$ by dimension}} \\
Task Progress           & .61  & & & \\
Instruction Consistency & .66  & & & \\
Object--Scene Cons.     & .59 & & & \\
Robot-Body Cons.        & .56 & & & \\
Physical Plausibility   & .51  & & & \\
Visual Quality          & .49  & & & \\
\bottomrule
\end{tabular}
\end{table}

\FloatBarrier

\subsection{Blinded Human Audit}
\label{app:humanvalid}

\begin{figure}[!h]
\centering
\includegraphics[width=0.82\linewidth]{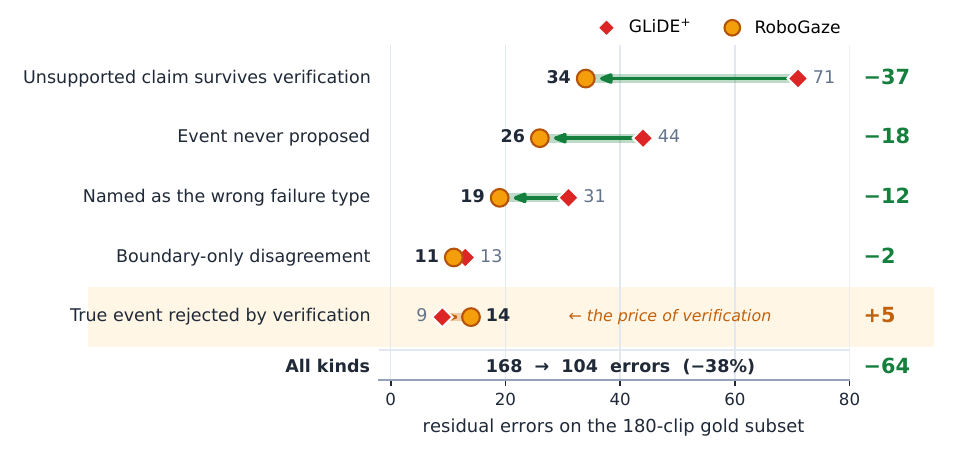}
\caption{Residual errors by kind on the gold subset, \gplus\ against
\textsc{RoboGaze}. Counts rather than shares, because the totals differ.
}
\label{fig:error_composition}
\end{figure}

The three senior annotators who produced the gold labels, blind to method identity, judged the Gemma4-31B predicted and
unrecovered events of \textsc{RoboGaze} and \gplus\ (B4b) on the $180$-clip gold subset
using the annotation rubric, and sorted every residual error into one of five
kinds (Figure~\ref{fig:error_composition}). Verification redistributes errors
rather than removing them evenly: it more than halves the unsupported claims that
reach the report, and the single kind \textsc{RoboGaze} produces more of than the
control is the true event its critic rejects. Unsupported physical claims remain
the largest kind for both methods, consistent with the per-dimension gap in
Figure~\ref{fig:agreement_gap}. Figure~\ref{fig:error_atlas} complements this
outcome-level view with representative cases grouped by the pipeline stage
that caused each error.

\begin{table}[!h]
\centering\small
\setlength{\tabcolsep}{4pt}
\caption{Human validation.}
\label{tab:humanvalid}
\begin{tabular}{llrr}
\toprule
Check & Condition & Value & $95\%$ CI \\
\midrule
\multirow{3}{*}{Human-audited event $F_1$}
 & B4b control        & 61.4 & [57.0, 65.7] \\
\rowcolor{rgaccent}
 & \textbf{\textsc{RoboGaze}} & \textbf{71.6} & [67.5, 75.4] \\
 & paired difference  & $+10.2$ & [6.3, 14.0] \\
\midrule
Pairwise preference & \textsc{RoboGaze} over B4b & 148/180 (82.2\%) & [75.9, 87.2] \\
\midrule
\multirow{2}{*}{Judge--human correlation}
 & text-only judge       & $r=.79$ & [.72, .85] \\
 & video-grounded judge  & $r=.83$ & [.77, .88] \\
\bottomrule
\end{tabular}
\end{table}

\FloatBarrier

\section{Implementation and Metrics}
\label{app:implmetrics}
\subsection{Implementation}
\label{app:impl}

Table~\ref{tab:model_snapshots} records the API snapshots and open-weight
checkpoints, Table~\ref{tab:visual_input} the per-backbone visual-input regime,
and Table~\ref{tab:controlconfig} the configuration of every method compared.
Because the input policy differs between backbones, method effects are
estimated within each backbone; the cross-backbone spread establishes breadth
only.

\begin{table}[!h]
\centering
\small
\caption{API access dates and checkpoint identifiers for the eight
evaluated backbones. For closed models the date is when the API endpoint was
first queried for our runs; providers may update an endpoint without notice. Open-source checkpoints were served using
\texttt{vLLM~0.20.2} on $4\!\times\!$NVIDIA L40 ($48$\,GB) with tensor
parallelism $=4$, batch size $1$ per request, and BF16 weights unless otherwise noted.}
\label{tab:model_snapshots}
\setlength{\tabcolsep}{3pt}
\begin{tabular}{>{\raggedright\arraybackslash}p{0.20\linewidth}>{\raggedright\arraybackslash}p{0.10\linewidth}>{\raggedright\arraybackslash}p{0.42\linewidth}>{\raggedright\arraybackslash}p{0.18\linewidth}}
\toprule
Backbone & Family & Access date / checkpoint & Context length \\
\midrule
Gemini 3.1 Pro      & closed & API, accessed $2026$-$04$-$15$ & $1$M tokens \\
GPT-5.5             & closed & API, accessed $2026$-$04$-$23$ & $1$M tokens \\
Gemini 3.1 Flash-Lite & closed & API, accessed $2026$-$04$-$15$ & $1$M tokens \\
Claude Sonnet 4.6   & closed & API, accessed $2026$-$04$-$15$ & $200$k tokens (API default) \\
Gemma4-31B          & open   & \texttt{google/gemma-4-31b-it}        & $256$k tokens \\
Qwen3.6-35B         & open   & \texttt{Qwen/Qwen3.6-35B-A3B}         & $262$k tokens \\
LLaVA-OV-2-8B       & open   & \texttt{lmms-lab/llava-ov-2-8b-it}    & $32$k tokens \\
InternVL3.5-38B     & open   & \texttt{OpenGVLab/InternVL3.5-38B}    & $32$k tokens \\
\bottomrule
\end{tabular}
\end{table}

\begin{table}[!h]
\centering
\small
\caption{Per-backbone visual-input interface, held fixed across all methods.
Closed APIs accept $16$ frames per call; open-weight interfaces accept up to
$32$ frames at $4$\,FPS. The latter is an interface cap. The default
\textsc{RoboGaze} observer uses $16$ frames per non-overlapping $4$\,s window;
the $32$-frame window is evaluated only in the window-length sweep. Multi-call
methods may therefore inspect more distinct frames per clip than a single
pass. Sensitivity to the regime is reported in
Appendix~\ref{app:robustness}.}
\label{tab:visual_input}
\begin{tabular}{lccc}
\toprule
Backbone & Mode & Effective rate & Per-call cap \\
\midrule
Gemini 3.1 Pro       & 16 uniformly sampled & $0.11$--$3.3$ FPS & 16 frames \\
Gemini 3.1 Flash-Lite & 16 uniformly sampled & $0.11$--$3.3$ FPS & 16 frames \\
GPT-5.5              & 16 uniformly sampled & $0.11$--$3.3$ FPS & 16 frames \\
Claude Sonnet 4.6    & 16 uniformly sampled & $0.11$--$3.3$ FPS & 16 frames \\
Gemma4-31B           & full video @ 4 FPS   & $0.21$--$4$ FPS         & 32 frames cap \\
Qwen3.6-35B          & full video @ 4 FPS   & $0.21$--$4$ FPS         & 32 frames cap \\
LLaVA-OV-2-8B        & full video @ 4 FPS   & $0.21$--$4$ FPS         & 32 frames cap \\
InternVL3.5-38B      & full video @ 4 FPS   & $0.21$--$4$ FPS         & 32 frames cap \\
\bottomrule
\end{tabular}
\end{table}

\begin{table}[!h]
\centering\scriptsize
\setlength{\tabcolsep}{3.2pt}
\caption{Configuration of every method compared. All share the backbone,
the backbone-specific visual-input interface, the matching rule and the scoring configuration; none is
fine-tuned.}
\label{tab:controlconfig}
\begin{tabular}{>{\raggedright\arraybackslash}p{0.05\linewidth}>{\raggedright\arraybackslash}p{0.19\linewidth}>{\raggedright\arraybackslash}p{0.24\linewidth}>{\raggedright\arraybackslash}p{0.22\linewidth}>{\raggedright\arraybackslash}p{0.18\linewidth}}
\toprule
& Method & Context supplied & Inference pattern & Verification \\
\midrule
--   & Vanilla & task instruction & 1 call, $T{=}0$ & none \\
--   & CoT & task instruction & 1 call, $T{=}0$, reasoning prefix & none \\
B1   & Full-context schema + CoT & task, initial scene, six-dimension, thirty-type schema, severity rubric, normal alternatives & 1 call, $T{=}0$ & none \\
B2   & B1 + self-consistency & as B1 & 9 samples, $T{=}0.7$; cluster at desc.\ sim.\ $0.7$ / IoU $0.5$; keep events with $\geq5$ supports & consensus only \\
B3   & Generic propose-and-verify & as B1 & propose, then verify each candidate; no subtasks or specialists & generic critic \\
B4a  & GLiDE + task/scene & task, initial scene & GLiDE memory, scanner, debate, frame re-inspection, temporal zoom, crop & debate \\
B4b  & B4a + schema/rubric & as B4a plus schema and rubric & as B4a & debate \\
B5   & \textsc{RoboGaze} w/o critic & task, scene, subtasks & routed specialists & none \\
\rowcolor{rgaccent}
B6   & \textbf{\textsc{RoboGaze}} & task, scene, subtasks & routed specialists over re-inspected windows & evidence critic + merge \\
\bottomrule
\end{tabular}
\end{table}

\paragraph{Settings.}
Verifier confidence threshold $0.50$ (swept in
Appendix~\ref{app:robustness}), temporal merge IoU $0.6$, router top-$k=3$
dimensions per subtask segment, non-overlapping $4$\,s windows (about five on a clip of mean duration, so
observation calls grow linearly with clip length; independent window calls are
issued in parallel; the choice of length is discussed below), temperature $0$
everywhere except B2, at most three retries on malformed output. A clip whose
output cannot be parsed after three retries is scored as an empty report rather
than dropped (on a failure-free clip this counts toward clean accuracy), which occurred for $0.4\%$ of clip--method pairs; retries enter the
compute accounting. Full prompt text is released with the code, each with a
version string and content hash recorded in the run manifest.

\paragraph{Window length.} The observer reads the clip in non-overlapping
windows of $4$\,s. The length follows from the per-call frame budget: at the
open-weight input rate of $4$\,FPS, a $4$\,s window fills one call's
$16$-frame budget, so every window is read with a frame every $0.25$\,s, and the
median annotated event ($1.4$\,s) spans several sampled frames inside a single
window. Shorter windows add observation calls without adding frame density;
longer windows must either spread the same $16$ frames more thinly or use a
larger call, up to the open-weight interface cap of $32$ frames.
Table~\ref{tab:window_length} reports \textsc{RoboGaze} on
Gemma4-31B with the window length varied and everything else fixed.
Halving the window to $2$\,s lowers $F_1$ to $55.8$, likely because more events
straddle window boundaries; doubling it to $8$\,s at the same $16$ frames halves the
frame density and lowers $F_1$ to $54.8$. Doubling both the window and the frames
recovers $57.0$, but requires $32$-frame calls, beyond the $16$-frame budget of the
closed-API backbones, so $4$\,s is the longest window that every backbone can read
at full density. Clean accuracy stays within $.85$--$.87$ throughout: the window
length changes what the specialists see, not what the critic rejects.

\begin{table}[!h]
\centering\small
\setlength{\tabcolsep}{5pt}
\caption{Window-length sweep, \textsc{RoboGaze} on Gemma4-31B, all $914$
clips; every other setting as in the main experiments. The $4$\,s row is the
default configuration of Table~\ref{tab:main_results}.}
\label{tab:window_length}
\begin{tabular}{lcccc}
\toprule
Window & Frames/window & $F_1\uparrow$ & $F_1{\times}$IoU$\uparrow$ & Clean$\uparrow$ \\
\midrule
$2$\,s & $8$  & $55.8$ & $33.5$ & $.85$ \\
\rowcolor{rgaccent}
$4$\,s (default) & $16$ & $\mathbf{57.3}$ & $\mathbf{36.1}$ & $.86$ \\
$8$\,s & $16$ & $54.8$ & $32.5$ & $\mathbf{.87}$ \\
$8$\,s & $32$ & $57.0$ & $35.5$ & $.86$ \\
\bottomrule
\end{tabular}
\end{table}

\FloatBarrier

\subsection{Metrics}
\label{app:metrics}

\paragraph{Judge.} Description similarity $S_{ij}$ is assigned by
Gemini~3.1~Pro (the snapshot in Table~\ref{tab:model_snapshots}) at temperature
$0$, from the two event descriptions only; malformed outputs are retried up to
three times. GPT-5.5 (Table~\ref{tab:model_snapshots}) serves as the alternate judge in
the robustness checks. Because the primary judge is also an evaluated backbone,
self-preference is tested directly by that swap (Appendix~\ref{app:robustness}).
The judge's similarity ratings correlate with human ratings of the same pairs at
$r=.79$ (Table~\ref{tab:humanvalid}).

\paragraph{Matching and scoring.}
Each predicted--reference pair is scored
$C_{ij} = S_{ij}\cdot\text{IoU}_t(\tau_i,\hat\tau_j)\cdot(1+0.25\,\mathbf{1}[\text{same dimension}])$,
so the dimension bonus breaks ties between otherwise similar candidates but cannot
create a match on its own. Pairwise temporal IoU is computed at the frame level. Pairs with
$\text{IoU}_t=0$ have $C_{ij}=0$ and are removed before matching; Hungarian
assignment runs in maximization mode and zero-cost pairs are dropped, so a pair
is matched only if $C_{ij}>0$. The score is multiplicative rather than additive
because a valid match must satisfy both semantic and temporal agreement.
Following GLiDE~\citep{zheng2026open}, with $N$ predicted and $M$ reference
events in a clip and matched pairs $\mathcal{A}$, description precision and recall
are $\sum_{(i,j)\in\mathcal{A}} S_{ij}/N$ and $\sum_{(i,j)\in\mathcal{A}} S_{ij}/M$,
and description $F_1$ is their harmonic mean; mIoU is the mean $\text{IoU}_t$ over
$\mathcal{A}$; and $F_1{\times}$IoU replaces $S_{ij}$ by $S_{ij}\,\text{IoU}_t$ in the same
precision--recall computation, so a matched event counts only as far as it both
describes and localizes the failure. All three are computed per clip and averaged
over clips; $F_1{\times}$IoU is therefore not the product of the reported $F_1$ and mIoU.
An unmatched prediction contributes to the precision denominator and an
unmatched reference to the recall denominator. An empty report on a clip with
reference events scores zero recall; on a clean clip it scores one for clean
accuracy and is excluded from the description metrics, whose denominators are
undefined there. Severe-event recall is computed over all severity-$4$/$5$
references including unmatched ones, so it cannot be inflated by answering only
on the events the evaluator found. Sev@1 is the percentage of matched events
whose predicted severity lies within $\pm1$ of the reference severity on the
$1$--$5$ scale.

Confidence intervals are bootstrap intervals over source conditions rather
than clips, since clips that share a source condition are not independent; $1{,}000$
resamples are drawn and the macro mean over clips recomputed in each. Method
comparisons are paired on the same clips and resampled jointly, so an interval on
a difference is an interval on the paired difference. No threshold is selected
separately for any backbone, generator or slice.

\FloatBarrier

\section{Complete Results and Robustness}
\label{app:results}

\subsection{Baselines and Measured Cost}
\label{app:controls}

Table~\ref{tab:controls} reports every method compared on two backbones;
Section~\ref{sec:mechanism} interprets the controls, and
Figure~\ref{fig:controls} plots them against cost. Two details are recorded
here. First, the shares of the CoT-to-\textsc{RoboGaze} distance closed by B1 and
B4b are $49\%$ and $68\%$ on Gemma4-31B and $49\%$ and $66\%$ on
Gemini~3.1~Pro, so the ordering of the controls does not depend on the backbone.
Second, on the Gemma deployment inference cost scales with clip length for
\textsc{RoboGaze} ($7.1$ calls and USD $0.138$ per clip below ten seconds,
$14.2$ calls and USD $0.309$ above thirty) and is flat for the single-pass
baselines on clips longer than $8$\,s, where the $32$-frame cap fixes their input.
\begin{table}[!h]
\centering\footnotesize
\setlength{\tabcolsep}{2.8pt}
\caption{Every method compared, on all $914$ clips. Sev-R counts all
severity-$4$/$5$ references, misses included. Budget columns are the Gemma4-31B
local deployment (4$\times$L40, BF16, TP$=$4, USD $1.20$ per GPU-hour), end-to-end
wall-clock time per clip (s/clip $\times$ $4$ GPUs $\times$ USD $1.20$), including encoding,
retries, tools and parsing; independent calls within a clip are issued
concurrently, each at batch size 1; Gemini API cost
per clip runs from USD $0.06$ (vanilla) to $0.52$ (B2) and $0.31$ for
\textsc{RoboGaze}.}
\label{tab:controls}
\begin{tabular}{@{}llrrrr rrr rrr@{}}
\toprule
& & \multicolumn{4}{c}{Gemma4-31B} & \multicolumn{3}{c}{Gemini 3.1 Pro} & \multicolumn{3}{c}{Budget (Gemma)} \\
\cmidrule(lr){3-6}\cmidrule(lr){7-9}\cmidrule(lr){10-12}
& Method & $F_1$ & $F_1{\times}$IoU & Sev-R & Clean & $F_1$ & $F_1{\times}$IoU & Clean & Calls & s/clip & USD \\
\midrule
--  & Vanilla                        & 16.4 &  5.1 & .61 & .04 & 24.1 &  9.2 & .09 & 1.0 &  21 & .028 \\
--  & CoT                            & 21.8 &  8.2 & .68 & .11 & 30.2 & 12.9 & .19 & 1.0 &  27 & .036 \\
B1  & Full-context schema + CoT    & 39.3 & 20.4 & .79 & .58 & 47.9 & 26.1 & .66 & 1.0 &  62 & .083 \\
B2  & B1 + self-consistency ($N{=}9$)& 43.4 & 22.9 & .83 & .64 & 51.6 & 28.8 & .71 & 9.0 & 168 & .224 \\
B3  & Generic propose-and-verify     & 42.1 & 22.1 & .81 & .66 & 50.2 & 27.9 & .72 & 6.0 & 149 & .199 \\
B4a & GLiDE + task/scene             & 43.6 & 23.2 & .84 & .65 & 51.8 & 29.0 & .71 & 9.0 & 141 & .188 \\
B4b & \gplus\ (B4a + schema)   & 45.9 & 24.8 & .86 & .69 & 54.1 & 30.1 & .74 &10.0 & 156 & .208 \\
B5  & \textsc{RoboGaze} w/o critic   & 44.1 & 23.4 & .94 & .21 & 52.6 & 28.6 & .29 & 7.6 & 118 & .157 \\
\rowcolor{rgaccent}
B6  & \textbf{\textsc{RoboGaze}}     & \textbf{57.3} & \textbf{36.1} & \textbf{.91} & \textbf{.86} & \textbf{66.4} & \textbf{42.4} & \textbf{.91} & 9.4 & 152 & .202 \\
\bottomrule
\end{tabular}
\end{table}

\subsection{Results by Slice}
\label{app:slices}

Task family is not independent of duration: coordinated and deformable
tasks are the longest in the benchmark, and their gain ($+12.8$) is close to what
the duration bands alone predict, so we do not read it as a separate effect.
Temporal mIoU by duration band is $.52/.45/.36/.30$ for B4b and
$.61/.57/.51/.47$ for \textsc{RoboGaze}. From the shortest band to the longest, B4b therefore
loses $15.4$ $F_1$ and $42\%$ of its temporal mIoU, against $9.0$ and $23\%$ for
\textsc{RoboGaze}. With sixteen uniformly sampled frames, $68\%$ of reference events in
clips above forty seconds fall entirely between two sampled frames, and more
inference over the same frames helps less and less: the advantage of B2 over B1
falls from $+4.8$ to $+2.0$ $F_1$ from short to long clips. The smallest gain
($+9.8$) falls on the one block that was quality-screened when first collected
(Appendix~\ref{app:filtering}).

\begin{figure}[!h]
\centering
\includegraphics[width=0.86\linewidth]{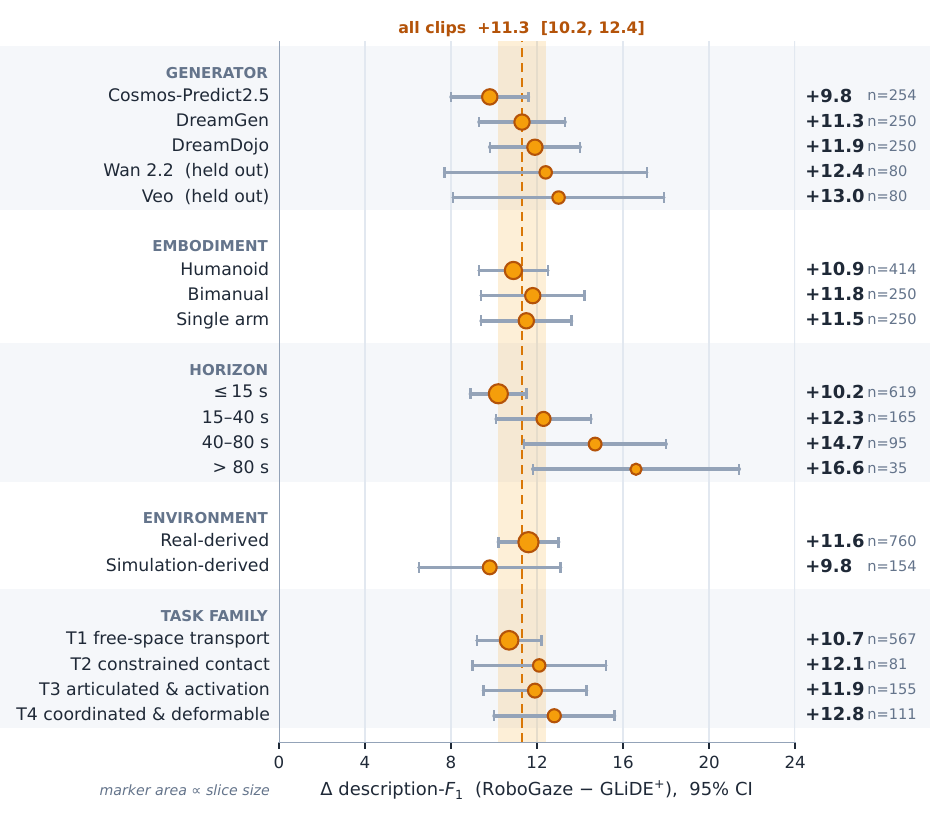}
\caption{Paired description-$F_1$ gain over B4b on every benchmark axis,
pooled over the eight backbones, with $95\%$ source-grouped bootstrap intervals.
All eighteen intervals exclude zero.}
\label{fig:slice_gains}
\end{figure}

\begin{table}[!h]
\centering\footnotesize
\setlength{\tabcolsep}{4pt}
\caption{Description $F_1$ by slice, pooled over the eight backbones.
$\Delta$ is the paired \textsc{RoboGaze}$-$B4b difference with a source-grouped
$95\%$ bootstrap interval.}
\label{tab:slices}
\begin{tabular}{llrrrrr}
\toprule
Axis & Slice & $n$ & CoT & B4b & \textbf{RG} & $\Delta$ \,[95\% CI] \\
\midrule
\multirow{5}{*}{Generator}
 & Cosmos-Predict2.5 & 254 & 25.8 & 49.4 & \textbf{59.2} & $+9.8$\,[8.0, 11.6] \\
 & DreamGen & 250 & 24.1 & 46.7 & \textbf{58.0} & $+11.3$\,[9.3, 13.3] \\
 & DreamDojo & 250 & 22.8 & 44.8 & \textbf{56.7} & $+11.9$\,[9.8, 14.0] \\
 & Wan 2.2 & 80 & 20.6 & 41.6 & \textbf{54.0} & $+12.4$\,[7.7, 17.1] \\
 & Veo & 80 & 19.9 & 40.4 & \textbf{53.4} & $+13.0$\,[8.1, 17.9] \\
\midrule
\multirow{3}{*}{Embodiment}
 & Humanoid & 414 & 24.3 & 47.3 & \textbf{58.2} & $+10.9$\,[9.3, 12.5] \\
 & Bimanual & 250 & 21.9 & 43.4 & \textbf{55.2} & $+11.8$\,[9.4, 14.2] \\
 & Single arm & 250 & 23.8 & 46.5 & \textbf{58.0} & $+11.5$\,[9.4, 13.6] \\
\midrule
\multirow{4}{*}{Duration}
 & $\leq 15$\,s & 619 & 25.2 & 48.4 & \textbf{58.6} & $+10.2$\,[8.9, 11.5] \\
 & $15$--$40$\,s & 165 & 22.4 & 44.6 & \textbf{56.9} & $+12.3$\,[10.1, 14.5] \\
 & $40$--$80$\,s & 95 & 17.9 & 37.6 & \textbf{52.3} & $+14.7$\,[11.4, 18.0] \\
 & $>80$\,s & 35 & 15.1 & 33.0 & \textbf{49.6} & $+16.6$\,[11.8, 21.4] \\
\midrule
\multirow{2}{*}{Environment}
 & Real-derived & 760 & 23.2 & 45.4 & \textbf{57.0} & $+11.6$\,[10.2, 13.0] \\
 & Simulation-derived & 154 & 25.4 & 49.2 & \textbf{59.0} & $+9.8$\,[6.5, 13.1] \\
\midrule
\multirow{4}{*}{Task family}
 & T1 free-space transport & 567 & 24.6 & 47.3 & \textbf{58.0} & $+10.7$\,[9.2, 12.2] \\
 & T2 constrained contact & 81 & 21.0 & 42.1 & \textbf{54.2} & $+12.1$\,[9.0, 15.2] \\
 & T3 articulated \& activation & 155 & 23.1 & 45.2 & \textbf{57.1} & $+11.9$\,[9.5, 14.3] \\
 & T4 coordinated \& deformable & 111 & 20.8 & 41.8 & \textbf{54.6} & $+12.8$\,[10.0, 15.6] \\
\midrule
\rowcolor{rgsummary}
\textbf{All} & \textbf{\textsc{RoboGazeBench}} & \textbf{914} & \textbf{23.5} & \textbf{46.0} & \textbf{57.3} & $\mathbf{+11.3}$\,[10.2, 12.4] \\
\bottomrule
\end{tabular}
\end{table}

\subsection{Complete Ablations, Per-Dimension Scores, and Judge-Free Metrics}
\label{app:ablations}

Scene memory matters more than task memory ($-5.7$ against $-4.6$ $F_1$),
and the two are not additive: removing both costs $6.8$ rather than $10.3$.
The critic is the only component whose removal raises severe-event recall, and
removing it together with the router costs barely more clean accuracy than
removing it alone ($.19$ against $.21$), so the false reports it suppresses are not
produced by routing.

\begin{figure}[!h]
\centering
\includegraphics[width=0.62\linewidth]{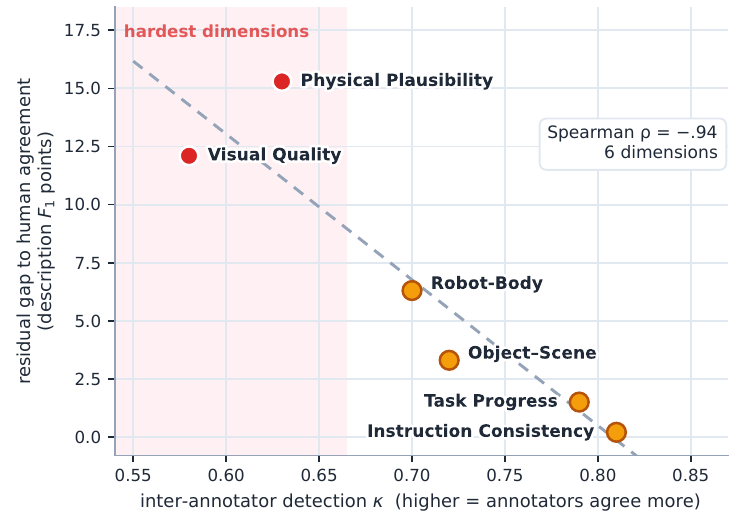}
\caption{Per dimension, the residual gap between human agreement and
\textsc{RoboGaze} (description $F_1$ points, Gemini~3.1~Pro, gold subset) against
inter-annotator detection $\kappa$ on that dimension. The dashed line is a
least-squares fit; the shaded region marks the two dimensions with the lowest
annotator agreement.}
\label{fig:agreement_gap}
\end{figure}

Per dimension, the residual gap to human agreement is inversely related to how
well annotators agree on that dimension (Figure~\ref{fig:agreement_gap},
Spearman $\rho=-.94$): $0.2$ $F_1$ on Instruction Consistency, where
$\kappa=.81$, against $15.3$ on Physical Plausibility, where $\kappa=.63$.
Without a language judge, the methods separate most on the $210$
failure-free clips: clean accuracy on Gemma4-31B is $.11$ for CoT, $.69$ for B4b
and $.86$ for \textsc{RoboGaze}, while severe-event recall, also judge-free, is
$.68$, $.86$ and $.91$ (Table~\ref{tab:controls}).
\begin{table}[!h]
\centering\small
\setlength{\tabcolsep}{5pt}
\caption{Complete component ablation, Gemma4-31B. $F_1$, $F_1{\times}$IoU and
Sev@1 (severity within $\pm1$ on matched events; Appendix~\ref{app:metrics}) are
percentages; Sev-R and Clean are fractions.}
\label{tab:ablations_full}
\begin{tabular}{lccccc}
\toprule
Configuration & $F_1\uparrow$ & $F_1{\times}$IoU$\uparrow$ & Sev@1$\uparrow$ & Sev-R$\uparrow$ & Clean$\uparrow$ \\
\midrule
\rowcolor{rgaccent}
\textbf{\textsc{RoboGaze}}       & \textbf{57.3} & \textbf{36.1} & \textbf{81.4} & \textbf{.91} & \textbf{.86} \\
\quad w/o task memory            & 52.7 & 32.0 & 77.2 & .88 & .82 \\
\quad w/o scene memory           & 51.6 & 31.0 & 76.1 & .87 & .80 \\
\quad w/o task \emph{and} scene memory & 50.5 & 29.8 & 75.0 & .86 & .79 \\
\quad w/o subtask segmentation   & 53.8 & 33.1 & 78.0 & .89 & .83 \\
\quad w/o candidate router       & 52.2 & 31.9 & 77.0 & .87 & .82 \\
\quad w/o dimension specialists  & 50.9 & 30.4 & 75.8 & .86 & .81 \\
\rowcolor{rgdrop}
\quad w/o critic verifier        & 44.1 & 23.4 & 71.2 & .94 & .21 \\
\quad w/o router \emph{and} critic & 42.0 & 21.8 & 69.5 & .93 & .19 \\
\bottomrule
\end{tabular}
\end{table}

\begin{table}[!h]
\begin{minipage}[t]{\linewidth}
\centering\small
\setlength{\tabcolsep}{4pt}
\caption{Per-dimension description $F_1$, $180$-clip gold subset,
Gemini~3.1~Pro.}
\label{tab:perdim}
\begin{tabular}{lrrrr}
\toprule
Dimension & CoT & B4b & \textbf{RG} & \emph{Human} \\
\midrule
Task Progress    & 34.1 & 63.8 & \textbf{76.4} & \emph{77.9} \\
Instruction Cons.& 36.9 & 67.1 & \textbf{77.0} & \emph{77.2} \\
Object--Scene    & 30.2 & 58.4 & \textbf{72.0} & \emph{75.3} \\
Robot-Body       & 27.4 & 54.0 & \textbf{67.3} & \emph{73.6} \\
\rowcolor{rgdrop}
Physical Plaus.  & 20.8 & 42.1 & \textbf{54.8} & \emph{70.1} \\
\rowcolor{rgdrop}
Visual Quality   & 22.6 & 44.6 & \textbf{56.3} & \emph{68.4} \\
\bottomrule
\end{tabular}
\end{minipage}
\end{table}

\subsection{Gold Subset}
\label{app:gold}

The human reference in Table~\ref{tab:main_results} is measured on the
$180$-clip gold subset rather than the full benchmark. Model scores on that
subset are within $0.9$ $F_1$ of their full-benchmark values
(Table~\ref{tab:gold}), so the two can be read against each other. The
gap-closure figures in Section~\ref{sec:main_results} are computed within this
subset, where model and human scores share the same clips: on Gemini~3.1~Pro,
\textsc{RoboGaze} recovers $(67.3-31.1)/(74.6-31.1)=83\%$ of the distance from
CoT to human agreement in description $F_1$, \gplus\ $55\%$, and
\textsc{RoboGaze} $99\%$ of it in clean accuracy.

\begin{table}[!h]
\centering\small
\setlength{\tabcolsep}{6pt}
\caption{Gold subset ($n{=}180$) against the full benchmark
($n{=}914$).}
\label{tab:gold}
\begin{tabular}{llrrrr}
\toprule
Backbone & Method & $F_1$ (gold) & $F_1$ (all) & Clean (gold) & Clean (all) \\
\midrule
\multirow{3}{*}{Gemini 3.1 Pro}
 & CoT & 31.1 & 30.2 & .20 & .19 \\
 & B4b & 55.0 & 54.1 & .75 & .74 \\
\rowcolor{rgaccent}
 & \textbf{\textsc{RoboGaze}} & \textbf{67.3} & \textbf{66.4} & \textbf{.92} & \textbf{.91} \\
\midrule
\multirow{3}{*}{Gemma4-31B}
 & CoT & 22.4 & 21.8 & .12 & .11 \\
 & B4b & 46.8 & 45.9 & .70 & .69 \\
\rowcolor{rgaccent}
 & \textbf{\textsc{RoboGaze}} & \textbf{58.2} & \textbf{57.3} & \textbf{.87} & \textbf{.86} \\
\midrule
\rowcolor{rgsummary}
\textit{Human agreement} & & \textit{74.6} & -- & \textit{.93} & -- \\
\bottomrule
\end{tabular}
\end{table}

\subsection{Offline Candidate Screening}
\label{app:screening}

For this study we rendered $40$ conditions six times each, under three
generators and two seeds, outside the $914$-clip benchmark; no screening
candidate is also a benchmark clip. Results are in Table~\ref{tab:screening}
(Section~\ref{sec:utility}); all evaluators run on Gemma4-31B, and per-clip costs there are measured on these candidates and are not
comparable with the benchmark-wide averages in Table~\ref{tab:controls}. Independent annotators, blind to every evaluator
output, labelled each candidate usable, unusable or uncertain as a \emph{visual}
demonstration; confirmed-usable prevalence is $118/240$ ($0.49$), so at $50\%$
retention a perfect ranker is bounded above by $0.98$ precision. We include two
scalar evaluators that cannot emit event-level reports (VideoScore2, a learned
judge, and the training-free RBench), scored through their
native outputs, because the question here is ranking rather than diagnosis. The
result concerns human-confirmed visual usability and not action executability,
simulator success or policy performance.

\begin{table}[!h]
\centering\small
\setlength{\tabcolsep}{5pt}
\caption{\textbf{Offline screening of generated demonstrations.} $40$
conditions $\times$ $6$ candidates rendered outside \textsc{RoboGazeBench}, ranked
by each evaluator; human annotators, blind to every evaluator, label each
candidate usable or not. P@50: usable precision in the retained half (random
$.49$, perfect $.98$); Bad@50: rate of severity-$4$/$5$ failures in the retained
half; Pick1: rate at which the top-ranked candidate is usable.}
\label{tab:screening}
\begin{tabular}{lrrrr}
\toprule
Evaluator & P@50$\uparrow$ & Bad@50$\downarrow$ & Pick1$\uparrow$ & USD/clip$\downarrow$ \\
\midrule
Random selection                & .49 & .20 & .49 & .00 \\
B1 Full-context schema + CoT  & .66 & .13 & .63 & .06 \\
B4b \gplus                      & .71 & .11 & .68 & .11 \\
VideoScore2~\citep{he2025videoscore2} & .65 & .14 & .61 & .04 \\
RBench (obj.\ + act.\ + phys.)~\citep{deng2026rethinking} & .62 & .15 & .58 & .05 \\
\rowcolor{rgaccent}
\textbf{\textsc{RoboGaze}}      & \textbf{.80} & \textbf{.05} & \textbf{.78} & .10 \\
\bottomrule
\end{tabular}
\end{table}
\FloatBarrier

\subsection{Robustness Checks}
\label{app:robustness}

\begin{figure}[!h]
\centering
\includegraphics[width=0.62\linewidth]{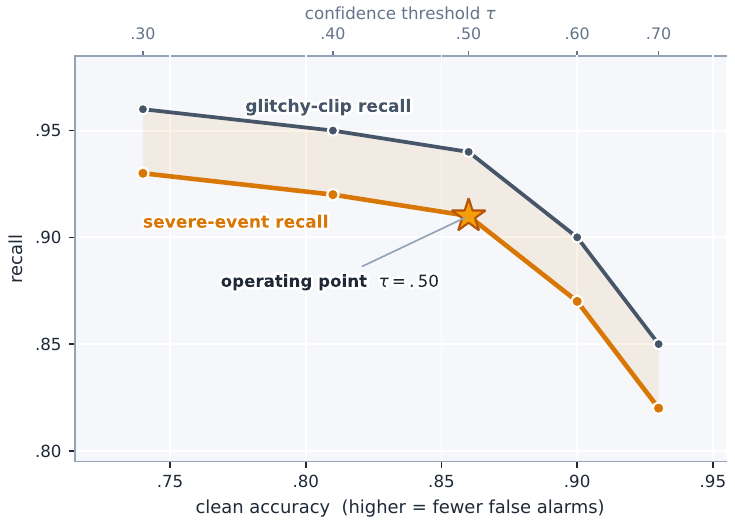}
\caption{Verifier operating curve, Gemma4-31B: recall against clean
accuracy as the confidence threshold varies over $.30$--$.70$. The starred point
is the operating threshold, $\tau=.50$, fixed on development data. Description
$F_1$ over the same range is $54.7$, $56.5$, $57.3$, $56.6$, $54.4$.
}
\label{fig:threshold}
\end{figure}

No check below changes the ordering of Table~\ref{tab:main_results}. The largest
effect on the \textsc{RoboGaze}$-$B4b margin is $0.8$ $F_1$, from halving the
frame budget; every other check moves it by at most $0.5$.
Figure~\ref{fig:threshold} gives the verifier operating curve; $0.50$ was fixed
on the development pool and the curve is reported so that a deployer can choose
a different point deliberately.
\begin{table}[!h]
\centering\footnotesize
\setlength{\tabcolsep}{3.6pt}
\caption{Robustness checks.}
\label{tab:robustness_index}
\begin{tabular}{>{\raggedright\arraybackslash}p{0.26\linewidth}>{\raggedright\arraybackslash}p{0.31\linewidth}>{\raggedright\arraybackslash}p{0.35\linewidth}}
\toprule
If this were true & Check & Result \\
\midrule
The judge favours its own model family & re-score all cells with GPT-5.5 & $\rho=.95$ over all $96$ backbone$\times$method$\times$metric cells, $24$ of which (clean accuracy) no judge affects, $\max|\Delta|=1.9$, mean $|\Delta|=0.8$ \\
The result needs \emph{some} LLM judge & BERTScore matching; judge-free metrics & backbone ranking $\rho=.93$, $61\%$ of the gap retained; clean accuracy and Sev-R use no judge (Table~\ref{tab:controls}) \\
The gain is explained by inference budget & B1 $\to$ B2, nine samples over identical evidence & $+4.1$ $F_1$ at $2.7\times$ cost, $23\%$ of the residual gap; $+4.8 \to +2.0$ from short to long clips \\
The gain is the schema & B1 and B4b hold the full schema & they close $49\%$ and $68\%$ of the distance but reach only $.58$ and $.69$ clean accuracy \\
The operating point was chosen post hoc & threshold sweep, fixed on development data & ordering stable over $.30$--$.70$ (Figure~\ref{fig:threshold}) \\
The matching rule favours \textsc{RoboGaze} & $\lambda_{\rm dim}\in\{0,.25,.5\}$; strict vs.\ loose & margin $11.0$/$11.4$/$11.4$; loose $10.9$; no ordering changes \\
Frame sampling drives the result & half, default and double the frame budget & RG $54.8$/$57.3$/$58.1$, B4b $44.2$/$45.9$/$46.4$; margin $10.6$/$11.4$/$11.7$ \\
\bottomrule
\end{tabular}
\end{table}

\FloatBarrier

\section{Qualitative Atlas, Release, and Ethics}
\label{app:atlas}

\begin{figure}[!h]
\centering
\includegraphics[width=\textwidth]{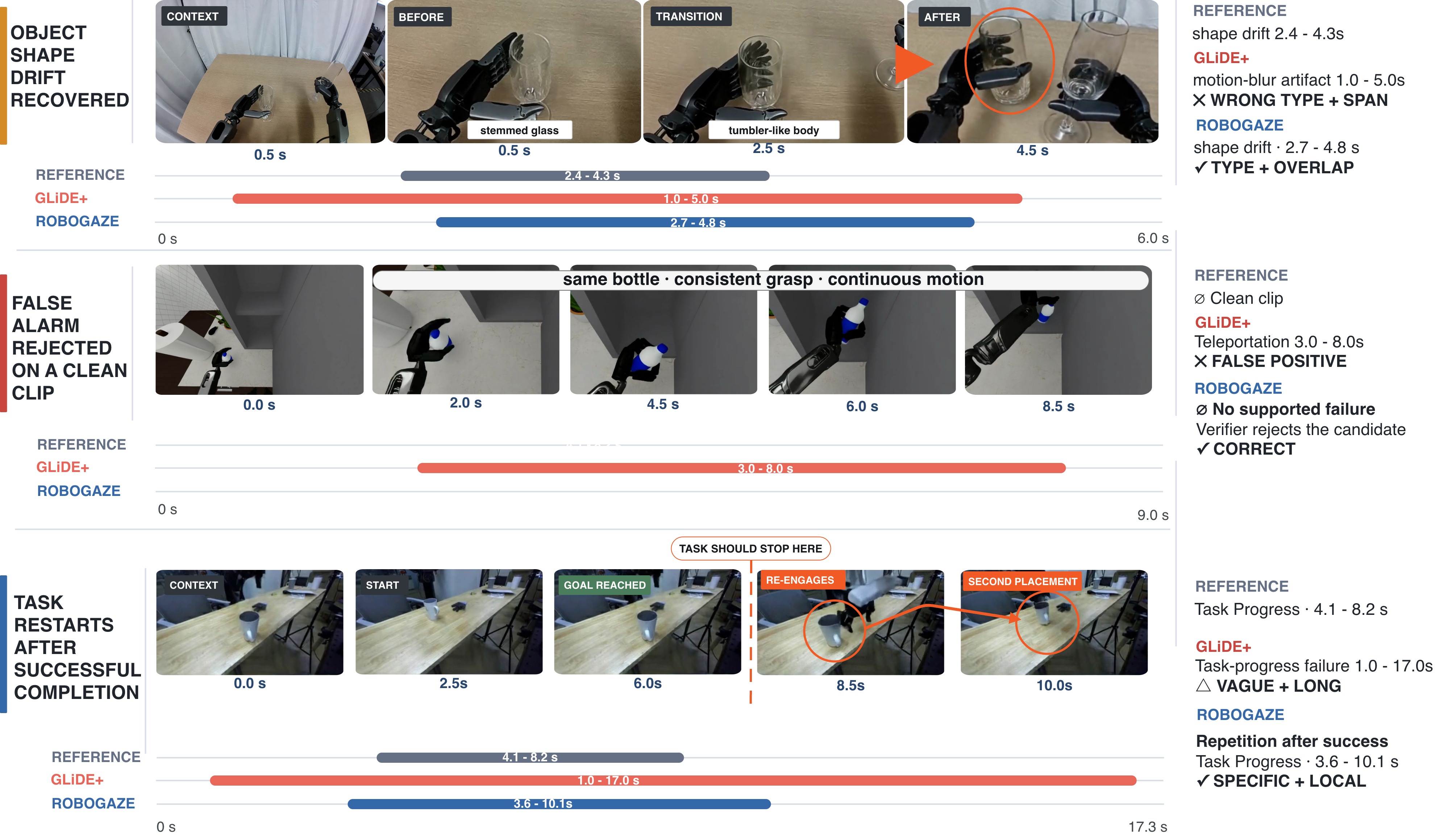}
\caption{\textbf{Where structured diagnosis pays off.} Three representative
single-view clips compare the human reference (gray), \gplus\ (red), and
\textsc{RoboGaze} (blue). (a) \gplus\ misclassifies an object distortion (shape drift) as motion blur,
whereas \textsc{RoboGaze} recovers the failure type and overlapping span.
(b) The verifier rejects an unsupported teleportation report on a clean clip.
(c) Both methods detect a post-completion task restart, but \textsc{RoboGaze}
provides a more specific diagnosis and tighter localization. Frames are
unmodified; complementary failure cases appear in
Figure~\ref{fig:error_atlas}.}
\label{fig:qualitative}
\end{figure}

\begin{figure}[p]
\centering
\includegraphics[width=\linewidth]{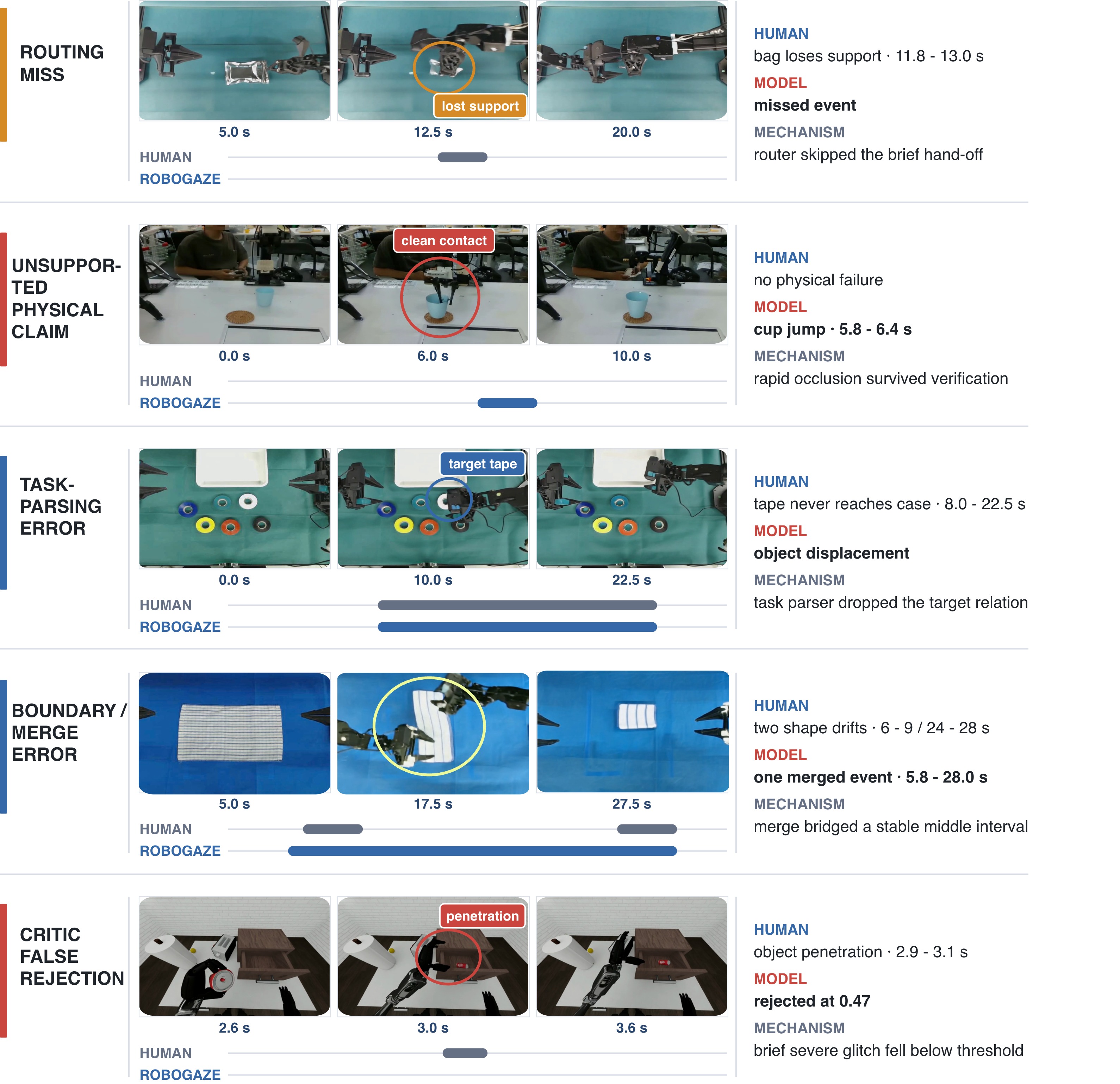}
\caption{\textbf{Where \textsc{RoboGaze} fails.} One case per error mechanism,
i.e.\ the pipeline stage that caused the error (a coding complementary to the
outcome kinds of Figure~\ref{fig:error_composition}), ordered by frequency, on five different
generator groups. Each row shows unmodified frames spanning the event, the human
reference and the predicted span, and the mechanism behind the error. Clip
identifiers, timestamps and crop coordinates are in the release.
Figure~\ref{fig:qualitative} shows the complementary successful cases.}
\label{fig:error_atlas}
\end{figure}

Every frame shown in this paper is an unmodified benchmark frame. Crops are
lossless and overlays are vector graphics; no frame has been upscaled, enhanced
or generated by an image model, and captions state whether a panel is raw,
cropped, annotated, plotted or schematic. The clip identifier, generator group, timestamp range and crop coordinates of
every displayed case are recorded in the release.

The release, made public upon acceptance, will contain all $914$ clips with their conditioning instructions and
initial observations, the manifest, per-annotator labels and adjudication
records, the annotation guide and interface, every prompt with its version and
hash, the evaluator and baseline code including all configurations, every
model output, the scoring code, and the analysis scripts that produce each table
and figure. Source datasets are redistributed only under their original licences.

\textsc{RoboGaze} reports visible failures in generated video. An empty report
means no sufficiently supported visible evidence of failure was found; it is not
a certificate of physical validity and must not be used as a safety
certification, as evidence of executability, or as a substitute for human
oversight where an incorrect diagnosis could propagate into a physical system.
The operating curve in Figure~\ref{fig:threshold} is reported so that a deployer
chooses a point rather than inheriting ours.

\end{document}